%% file: capgen.tex
\documentclass{article}

\usepackage{times}
\usepackage{graphicx} 
\usepackage{subfigure} 
\usepackage{amsmath}
\usepackage{amsfonts}

\usepackage{stfloats}

\usepackage{url}

\usepackage{natbib}

\usepackage{algorithm}
\usepackage{algorithmic}

\usepackage{hyperref}

\usepackage[accepted]{arXiv}

\graphicspath{ {./figures/} }

\newcommand{\bmx}[0]{\begin{bmatrix}}
\newcommand{\emx}[0]{\end{bmatrix}}

\newcommand{\vect}[1]{\mathbf{#1}}

\newcommand{\matr}[1]{\mathbf{#1}}

\newcommand{\va}[0]{\vect{a}}

\newcommand{\vc}[0]{\vect{c}}

\newcommand{\vh}[0]{\vect{h}}

\newcommand{\vn}[0]{\vect{n}}
\newcommand{\vz}[0]{\vect{z}}

\newcommand{\vf}[0]{\vect{f}}
\newcommand{\vi}[0]{\vect{i}}
\newcommand{\vo}[0]{\vect{o}}
\newcommand{\vy}[0]{\vect{y}}
\newcommand{\vg}[0]{\vect{g}}

\newcommand{\vL}[0]{\vect{L}}

\newcommand{\vE}[0]{\matr{E}}

\newcommand{\RR}[0]{\mathbb{R}}

\newcommand{\sigmoid}{\sigma}

\newcommand{\ts}{\rule{0pt}{2.6ex}}       
\newcommand{\specialcell}[2][c]{%
  \begin{tabular}[#1]{@{}c@{}}#2\end{tabular}}

\icmltitlerunning{Neural Image Caption Generation with Visual Attention}

\begin{document} 

\twocolumn[
\icmltitle{Show, Attend and Tell: Neural Image Caption\\Generation with Visual Attention}

\icmlauthor{Kelvin Xu}{kelvin.xu@umontreal.ca}
\icmlauthor{Jimmy Lei Ba}{jimmy@psi.utoronto.ca}
\icmlauthor{Ryan Kiros}{rkiros@cs.toronto.edu}
\icmlauthor{Kyunghyun Cho}{kyunghyun.cho@umontreal.ca}
\icmlauthor{Aaron Courville}{aaron.courville@umontreal.ca}
\icmlauthor{Ruslan Salakhutdinov}{rsalakhu@cs.toronto.edu}
\icmlauthor{Richard S. Zemel}{zemel@cs.toronto.edu}
\icmlauthor{Yoshua Bengio}{find-me@the.web}


\vskip 0.3in
]

\input{main.tex}

%
%

\small
\bibliography{capgen}
\bibliographystyle{icml2015}
\newpage
\input{appendix.tex}

\end{document}

%% file: main.tex
\begin{abstract} 
Inspired by recent work in machine translation and object detection, we
introduce an attention based model that automatically learns to describe the
content of images. We describe how we can train this model in a deterministic
manner using standard backpropagation techniques and stochastically by
maximizing a variational lower bound. We also show through visualization how
the model is able to automatically learn to fix its gaze on salient objects
while generating the corresponding words in the output sequence.
We validate the use of attention with state-of-the-art performance on three
benchmark datasets: Flickr8k, Flickr30k and MS COCO.
\end{abstract} 

\section{Introduction}
Automatically generating captions of an image is a task very close to the heart
of scene understanding --- one of the primary goals of computer vision.  Not only
must caption generation models be powerful enough to solve the computer vision
challenges of determining which objects are in an image, but they must also be
capable of capturing and expressing their relationships in a natural language.
For this reason, caption generation has long been viewed as a difficult
problem.  It is a very important challenge for machine learning
algorithms, as it amounts to mimicking the remarkable human ability to compress
huge amounts of salient visual infomation into descriptive language.

Despite the challenging nature of this task, there has been a
recent surge of research interest in attacking the image caption generation
problem. Aided by advances in training neural networks \citep{Krizhevsky2012}
and large classification datasets \citep{Imagenet14}, recent work has
significantly improved the quality of caption generation using a combination of
convolutional neural networks (convnets) to obtain vectorial representation of images and
recurrent neural networks to decode those representations into natural language
sentences (see Sec.~\ref{section:background}). 

\begin{figure}[tp]
    \label{figure:model_diagram}
    \centering
        \caption{Our model learns a words/image alignment. The visualized
        attentional maps (3) are explained in section \ref{section:model} \& \ref{section:viz}}
        \vspace{3mm}
        \includegraphics[width=\columnwidth]{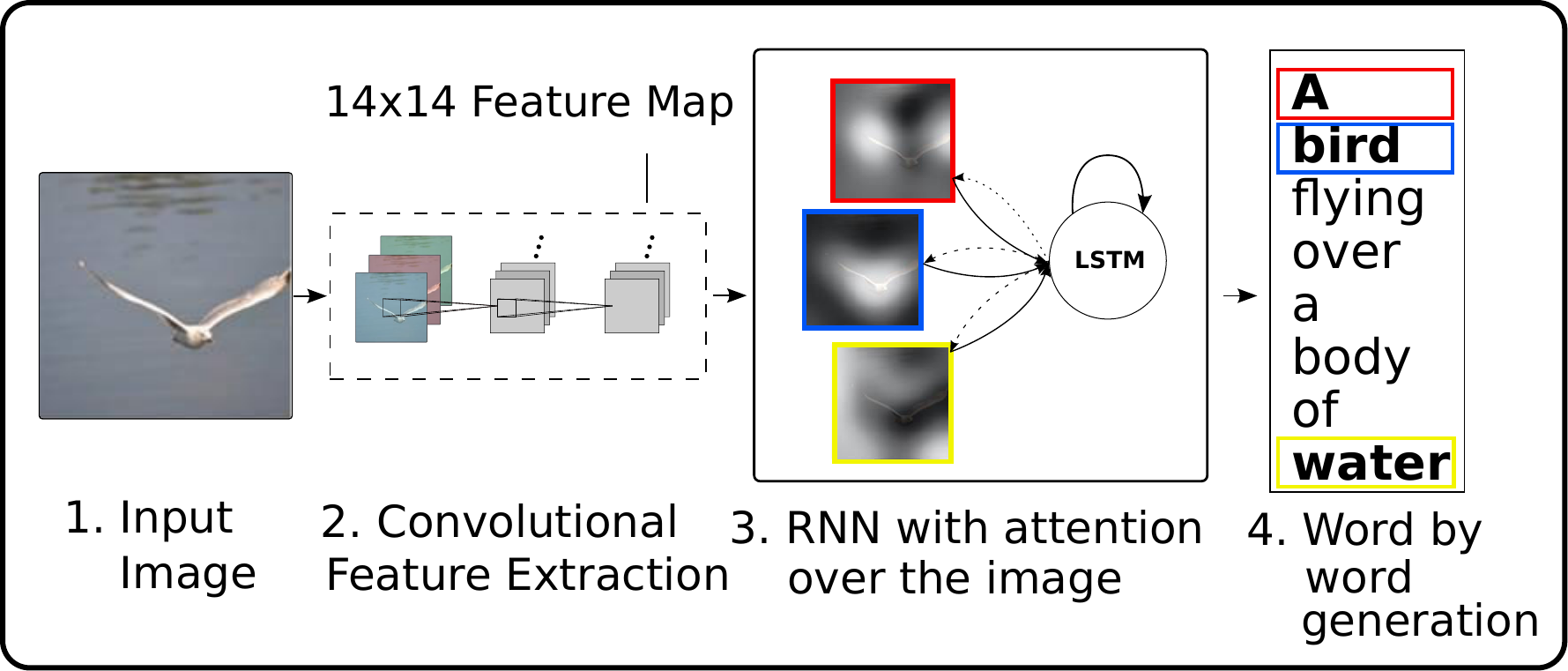}
        \vspace{-6mm}
\end{figure}
\begin{figure*}[!tp]
    \label{figure:attention_diagram}
    \centering
        \caption{Attention over time. As the model generates each word, its attention changes to reflect the relevant parts of the image. ``soft''
         (top row) vs ``hard'' (bottom row) attention. (Note that both models generated the same captions in this example.) }
        \includegraphics[width=6.5in]{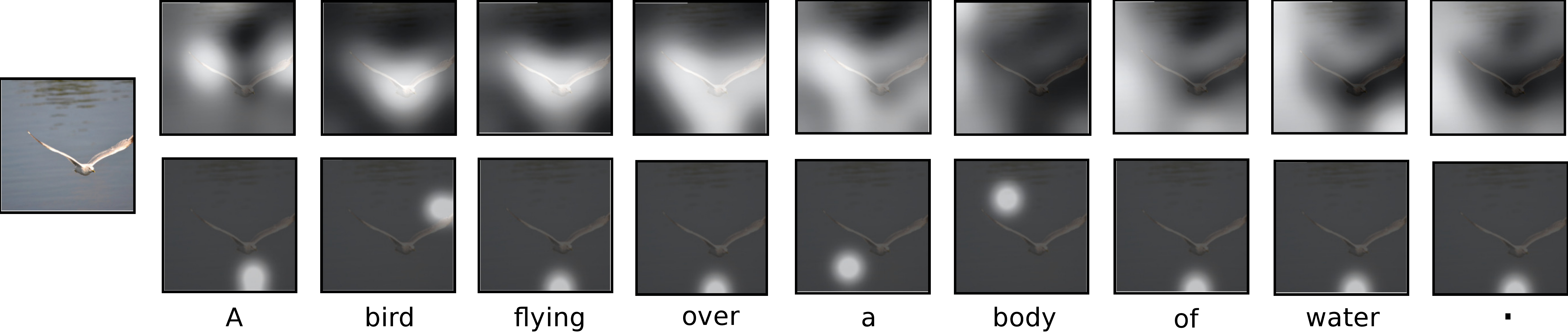}
        \vspace{-5mm}
\end{figure*}
\begin{figure*}[!tp]           
    \label{figure:alignments}          
    \centering         
        \caption{Examples of attending to the correct object (\textit{white} indicates the attended regions, 
        \textit{underlines} indicated the corresponding word)}                
        \includegraphics[width=0.87\textwidth]{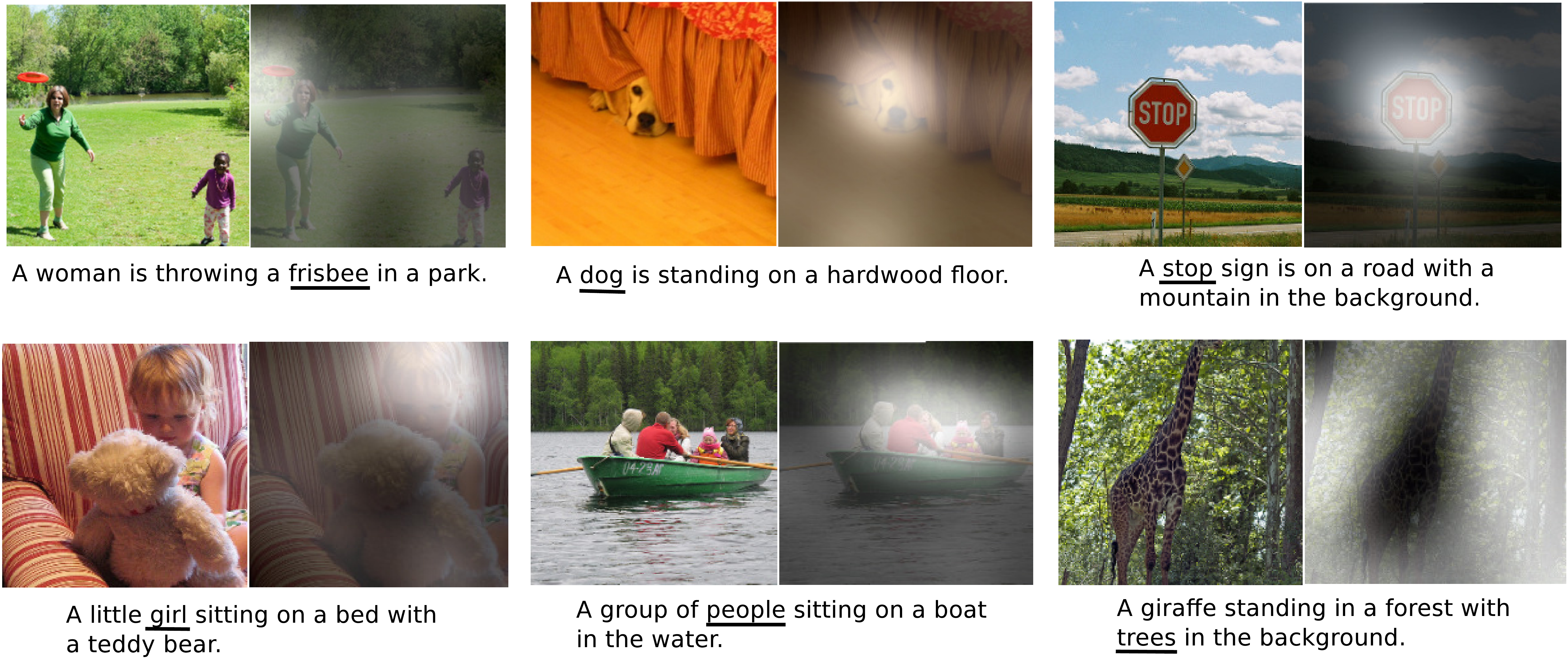}               
        \vspace{-5mm}
\end{figure*}

One of the most curious facets of the human visual system is the presence of
attention \citep{Rensink2000,Corbetta2002}. Rather than compress an entire
image into a static representation, attention allows for salient features to
dynamically come to the forefront as needed. This is especially important when
there is a lot of clutter in an image. Using representations (such as those
from the top layer of a convnet) that distill information in image down to the
most salient objects is one effective solution that has been widely adopted in
previous work. Unfortunately, this has one potential drawback of losing
information which could be useful for richer, more descriptive captions. Using
more low-level representation can help preserve this information. However
working with these features necessitates a powerful mechanism to steer the
model to information important to the task at hand.

In this paper, we describe approaches to caption generation that attempt to
incorporate a form of attention with two variants: a ``hard'' attention
mechanism and a ``soft'' attention mechanism. We also show how one advantage of
including attention is the ability to visualize what the model ``sees''.
Encouraged by recent advances in caption generation and inspired by recent
success in employing attention in machine translation \citep{Bahdanau2014} and
object recognition \citep{Ba2014,Mnih2014}, we investigate models that can
attend to salient part of an image while generating its caption. 

The contributions of this paper are the following:
\vspace{- 1mm}
\begin{itemize}
    \vspace{-2mm}
    \item We introduce two attention-based image caption generators under
    a common framework (Sec.~\ref{section:model}): 1) a ``soft'' deterministic attention mechanism trainable by standard 
    back-propagation methods and 2) a ``hard'' stochastic attention mechanism trainable by
    maximizing an approximate variational lower bound or equivalently by
    REINFORCE~\citep{Williams92}.
     \vspace{-2mm}
    \item We show how we can gain insight and interpret the results of 
    this framework by visualizing ``where'' and ``what'' the attention focused on. 
    (see Sec.~\ref{section:viz}) 
    \vspace{-2mm}
    \item  Finally, we quantitatively validate the usefulness of attention in
    caption generation with state of the art performance
    (Sec.~\ref{section:results}) on three benchmark datasets: Flickr8k
    \citep{Hodosh2013} , Flickr30k \citep{Young2014} and the MS COCO dataset
    \citep{Lin2014}.
\end{itemize}
\section{Related Work}
\label{section:background}

In this section we provide relevant background on previous work on
image caption generation and attention. 
Recently, several methods have been proposed for generating image descriptions.
Many of these methods are based on recurrent neural networks and inspired
by the successful use of sequence to sequence training with neural networks for
machine translation~\citep{Cho2014,Bahdanau2014,Sutskever2014}. One
major reason image caption generation is well suited to the encoder-decoder framework
\citep{Cho2014} of machine translation is because it is analogous
to ``translating'' an image to a sentence.

The first approach to use neural networks for caption generation was
\citet{Kiros2014a}, who proposed a multimodal log-bilinear model that was
biased by features from the image. This work was later followed by
\citet{Kiros2014b} whose method was designed to explicitly allow a natural way
of doing both ranking and generation.  \citet{Mao2014} took a similar approach
to generation but replaced a feed-forward neural language model with a
recurrent one. Both \citet{Vinyals2014} and \citet{Donahue2014} use LSTM RNNs
for their models.  Unlike \citet{Kiros2014a} and \citet{Mao2014} whose models
see the image at each time step of the output word sequence,
\citet{Vinyals2014} only show the image to the RNN at the beginning. Along
with images, \citet{Donahue2014} also apply LSTMs to videos, allowing
their model to generate video descriptions.

All of these works represent images as a single feature vector from the top
layer of a pre-trained convolutional network. \citet{Karpathy2014} instead
proposed to learn a joint embedding space for ranking and generation whose
model learns to score sentence and image similarity as a function of R-CNN
object detections with outputs of a bidirectional RNN. 
\citet{Fang2014} proposed a three-step pipeline for
generation by incorporating
object detections. 
Their model first learn detectors for several visual concepts
based on a multi-instance learning framework. A language model trained on
captions was then applied to the detector outputs, followed by rescoring from a
joint image-text embedding space. Unlike these models, our proposed attention
framework does not explicitly use object detectors but instead learns latent
alignments from scratch. This allows our model to go beyond ``objectness'' and
learn to attend to abstract concepts.

Prior to the use of neural networks for generating captions, two main
approaches were dominant. The first involved generating caption templates which
were filled in based on the results of object detections and attribute
discovery (\citet{Kulkarni2013}, \citet{Li2011},
\citet{Yang2011}, \citet{Mitchell2012}, \citet{Elliott2013}). The second
approach was based on first retrieving similar captioned images from a large
database then modifying these retrieved captions to fit the query
\citep{Kuznetsova2012,Kuznetsova2014}.  These approaches typically involved an
intermediate ``generalization'' step to remove the specifics of a caption that
are only relevant to the retrieved image, such as the name of a city. Both of
these approaches have since fallen out of favour to the now dominant neural
network methods.

There has been a long line of previous work incorporating attention into neural
networks for vision related tasks. Some that share the same spirit as our work
include \citet{Larochelle2010,Denil2012,Tang2014}. In particular however, our
work directly extends the work of \citet{Bahdanau2014,Mnih2014,Ba2014}.

\section{Image Caption Generation with Attention Mechanism}

\subsection{Model Details}
\label{section:model}
In this section, we describe the two variants of our attention-based model by
first describing their common framework. The main difference is the definition
of the $\phi$ function which we describe in detail in Section
\ref{sec:det_sto}. We denote vectors with bolded font and matrices with
capital letters. In our description below, we suppress bias terms for
readability.

\subsubsection{Encoder: Convolutional Features}

Our model takes a single raw image and generates a caption $\vy$
encoded as a sequence of 1-of-$K$ encoded words.
\[
    y = \left\{\vy_1, \ldots, \vy_{C} \right\},\mbox{ } \vy_i \in \RR^{K}
\]
where $K$ is the size of the vocabulary and $C$ is the length of the caption.

We use a convolutional neural network in order to extract a set of feature
vectors which we refer to as annotation vectors.  The extractor produces $L$
vectors, each of which is a D-dimensional representation corresponding to a
part of the image. 
\begin{align*}
    a = \left\{\va_1, \ldots, \va_L \right\},\mbox{ } \va_i \in \RR^{D}
 \end{align*}
In order to obtain a correspondence between the feature vectors and portions of
the 2-D image, we extract features from a lower convolutional layer unlike
previous work which instead used a fully connected layer. This allows the
decoder to selectively focus on certain parts of an image by selecting a subset
of all the feature vectors.
\begin{figure}[tp]
\vskip 0.2in
\begin{center}
\centerline{\includegraphics[width=\columnwidth]{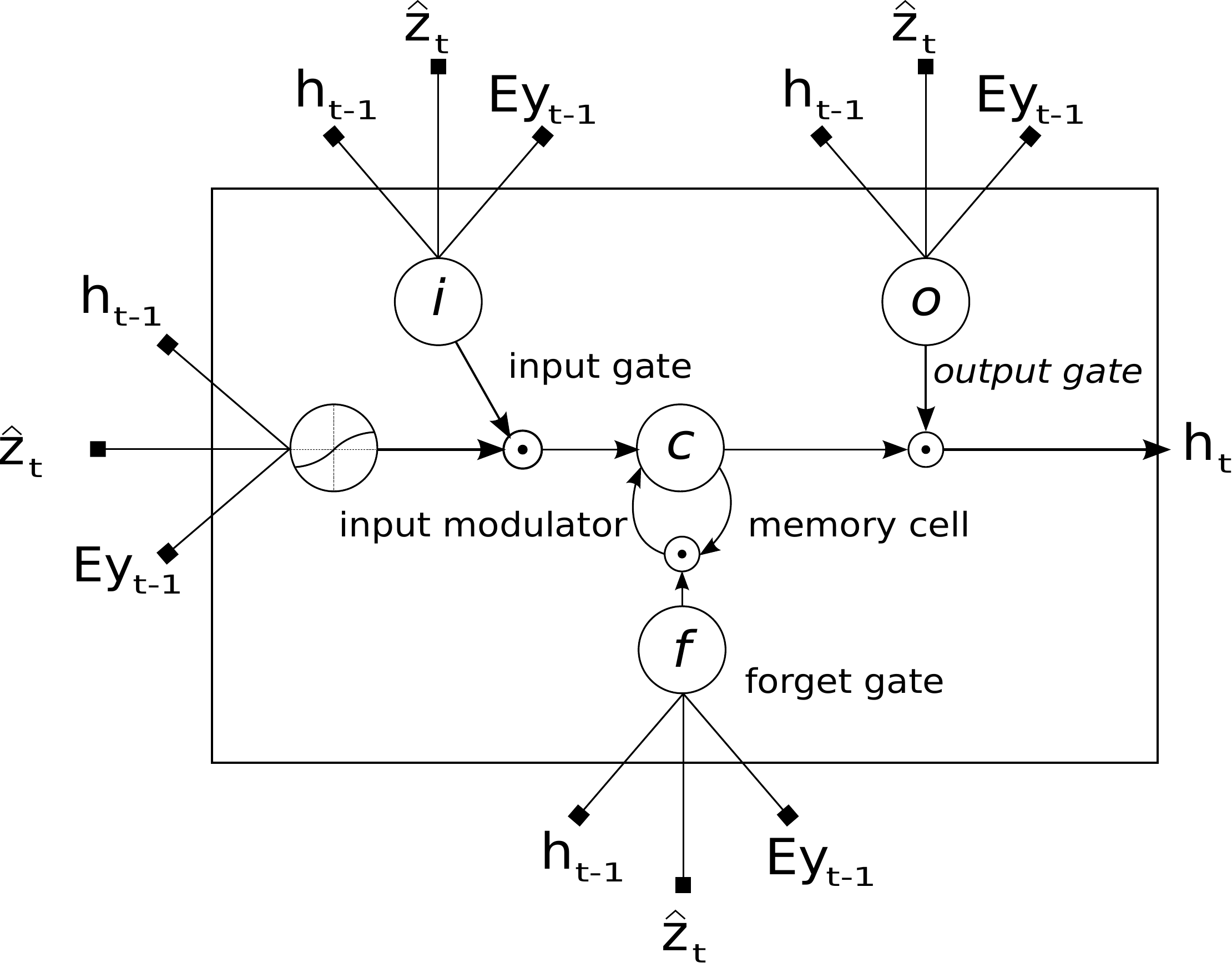}}
\caption{A LSTM cell, lines with bolded squares imply projections with a
learnt weight vector. Each cell learns how to weigh
its input components (input gate), while learning how to modulate that
contribution to the memory (input modulator). It also learns 
weights which erase the memory cell (forget gate), and weights 
which control how this memory should be emitted (output gate).}
\label{figure:conditional_lstm}
\end{center}
\vskip -0.3 in
\end{figure} 
\subsubsection{Decoder: Long Short-Term Memory Network}

We use a long short-term memory (LSTM)
network~\citep{Hochreiter+Schmidhuber-1997} that produces a caption by
generating one word at every time step conditioned on a context vector, the
previous hidden state and the previously generated words. Our implementation of
LSTM closely follows the one used in \citet{Zaremba2014} (see Fig.~\ref{figure:conditional_lstm}).  Using $T_{s,t}:
\RR^{s} \rightarrow \RR^{t}$ to denote a simple affine transformation with
parameters that are learned,

\begin{align}
\label{eq:lstm_gates}
\begin{pmatrix}
\vi_t \\
\vf_t \\ 
\vo_t \\
\vg_t \\
\end{pmatrix} =
&
\begin{pmatrix}
\sigmoid \\
\sigmoid \\ 
\sigmoid \\
\tanh \\
\end{pmatrix}
T_{D+m+n, n}
\begin{pmatrix}
\vE\vy_{t-1}\\
\vh_{t-1}\\
\hat{\vz_t}\\
\end{pmatrix}
\\
\label{eq:lstm_memory}
\vc_t &= \vf_t \odot \vc_{t-1} + \vi_t \odot \vg_t \\
\label{eq:lstm_hidden}
\vh_t &= \vo_t \odot \tanh (\vc_{t}). 
\end{align} 
Here, $\vi_t$, $\vf_t$, $\vc_t$, $\vo_t$, $\vh_t$ are the input, forget, memory, output
and hidden state of the LSTM, respectively. The vector $\hat{\vz} \in \RR^{D}$ is the context
vector, capturing the visual information associated with a particular
input location, as explained below. $\vE\in\RR^{m\times K}$ is an embedding matrix. Let $m$ and $n$
denote the embedding and LSTM dimensionality respectively and $\sigma$ and
$\odot$ be the logistic sigmoid activation and element-wise multiplication
respectively. 

In simple terms, the context vector $\hat{\vz}_t$ (equations~\eqref{eq:lstm_gates}--\eqref{eq:lstm_hidden}) 
is a dynamic representation of the relevant part of the image input at time $t$. 
We define a mechanism $\phi$ that computes $\hat{\vz}_t$ from the annotation vectors $\va_i, i=1,\ldots,L$
corresponding to the features extracted at different image locations. For
each location $i$, the
mechanism generates a positive weight $\alpha_i$ which can
be interpreted either as the probability that location $i$ is the right place to focus
for producing the next word (the ``hard'' but stochastic attention mechanism), 
or as the relative importance to give to location $i$ in blending the $a_i$'s together.
The weight $\alpha_i$ of each annotation vector $a_i$ is computed by an
\emph{attention model} $f_{\mbox{att}}$ for which we use a multilayer perceptron conditioned on the previous hidden state $h_{t-1}$. 
The soft version of this attention mechanism was introduced by~\citet{Bahdanau2014}.
For emphasis, we note that the hidden state varies as the output RNN advances in
its output sequence: ``where'' the network looks next depends on the sequence of words that has already been
generated.
\begin{align}
    e_{ti} =& f_{\mbox{att}} (\va_i, \vh_{t-1}) \\
    \label{eq:alpha}
    \alpha_{ti} =& \frac{\exp(e_{ti})}{\sum_{k=1}^L \exp(e_{tk})}.
\end{align}
Once the weights (which sum to one) are computed, the context vector $\hat{z}_t$
is computed by
\begin{align}
\label{eq:context}
    \hat{\vz}_t = \phi\left( \left\{ \va_i \right\}, \left\{ \alpha_i \right\}
    \right),
\end{align}
where $\phi$ is a function that returns a single vector given the set of
annotation vectors and their corresponding weights. The details of $\phi$ function
are discussed in Sec.~\ref{sec:det_sto}.

The initial memory state and hidden state of the LSTM are predicted by an
average of the annotation vectors fed through two separate MLPs
($\text{init,c}$ and $\text{init,h}$):
\begin{align} 
    \vc_0 = f_{\text{init,c}} (\frac{1}{L} \sum_i^L \va_i) \nonumber \\
    \vh_0 = f_{\text{init,h}} (\frac{1}{L} \sum_i^L \va_i) \nonumber 
\end{align}

In this work, we use a deep output layer~\citep{Pascanu2014} to compute the
output word probability given the LSTM state, the context vector and the
previous word:
\begin{gather}
\label{eq:p-out}
    p(\vy_t | \va, \vy_1^{t-1}) \propto \exp(\vL_o(\vE\vy_{t-1} + \vL_h\vh_t+ \vL_z \hat{\vz}_t))
\end{gather}

Where $\vL_o\in\RR^{K\times m}$, $\vL_h\in\RR^{m\times n}$, $\vL_z\in\RR^{m\times D}$, and
$\vE$ are learned parameters initialized randomly.

\section{Learning Stochastic ``Hard'' vs Deterministic ``Soft'' Attention}
\label{sec:det_sto}

In this section we discuss two alternative mechanisms for the attention
model $f_{\mbox{att}}$: stochastic attention and deterministic attention.
\begin{figure*}[!tp]
    \label{fig:second}
        \caption{Examples of mistakes where we can use attention to gain intuition into what the model saw.}
        \includegraphics[width=1.02\textwidth]{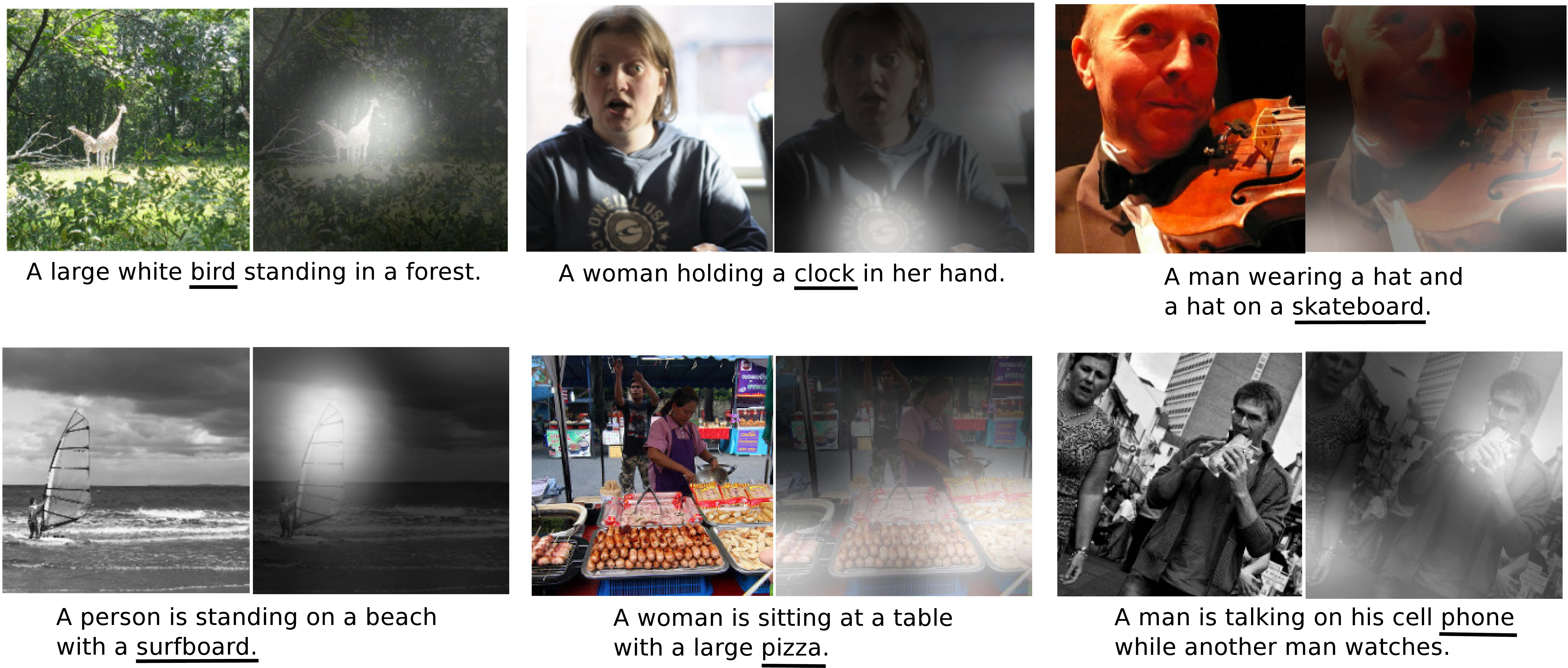}
   \label{fig:subfigures}
\end{figure*}
\subsection{Stochastic ``Hard'' Attention}
\label{sec:sto_attn}
We represent the location variable $s_t$ as where the model decides to focus attention
when generating the $t^{th}$ word. $s_{t,i}$ is an indicator one-hot variable which is set to 1
if the $i$-th location (out of $L$)
is the one used to extract visual features. By treating 
the attention locations as intermediate latent variables, we can 
assign a multinoulli distribution parametrized by $\{\alpha_i\}$,
and view $\hat{z}_t$ as a random variable:
\begin{align}
\label{eq:s_dist}
    p(&s_{t,i} = 1 \mid s_{j<t}, \va ) = \alpha_{t,i} \\
\label{eq:hard_context}
    \hat{\vz}_t &= \sum_i {s}_{t,i}\va_{i}.
\end{align}
We define a new objective function $L_s$ that is a variational lower bound on the marginal log-likelihood
$\log p( \vy \mid \va)$ of observing the sequence of words $\vy$ given image features $\va$.  
The learning algorithm for the parameters $W$ of the models can be derived by 
directly optimizing $L_s$:
\begin{align} 
    L_s  = & \sum_s p(s \mid \va) \log p(\vy \mid s, \va) \nonumber \\
     \leq & \log \sum_{s}p(s \mid \va)p( \vy \mid s, \va) \nonumber \\
     = & \log p( \vy \mid \va) \label{eq:mll}  
\end{align}
\begin{multline}
\label{eq:lb_gradient}
 \frac{\partial L_s}{\partial W} =  \sum_s p(s \mid \va) \bigg[ \frac{\partial \log p(\vy \mid s, \va)}{\partial W} + \\
                                \log p(\vy \mid s, \va) \frac{\partial \log p(s \mid \va)}{\partial W} \bigg].
\end{multline}
Equation \ref{eq:lb_gradient} suggests a Monte Carlo based sampling
approximation of the gradient with respect to the model parameters. This can be
done by sampling the location $s_t$ from a multinouilli distribution defined by
Equation \ref{eq:s_dist}.
\begin{gather}
\tilde{s_t} \sim \text{Multinoulli}_L(\{\alpha_i\}) \nonumber 
\end{gather}
\begin{multline}
 \frac{\partial L_s}{\partial W} \approx \frac{1}{N} \sum_{n=1}^{N} \bigg[ \frac{\partial \log p(\vy \mid \tilde{s}^n, \va)}{\partial W} +  \\
                                \log p(\vy \mid \tilde{s}^n, \va) \frac{\partial \log p(\tilde{s}^n \mid \va)}{\partial W} \bigg]
\end{multline} 
A moving average baseline is used to reduce the variance in the
Monte Carlo estimator of the gradient, following~\citet{Weaver+Tao-UAI2001}. 
Similar, but more complicated variance
reduction techniques have previously been used by \citet{Mnih2014} and
\citet{Ba2014}. Upon seeing the $k^{th}$ mini-batch, the moving average baseline 
is estimated as an accumulated sum of the previous log likelihoods with exponential decay:
\begin{align}
b_k = 0.9 \times b_{k-1} + 0.1 \times \log p(\vy \mid \tilde{s}_k , \va) \nonumber 
\end{align}
To further reduce the estimator variance, an entropy term on the multinouilli
distribution $H[s]$ is added. Also, with probability 0.5 for a given image, we
set the sampled attention location $\tilde{s}$ to its expected value $\alpha$.
Both techniques improve the robustness of the stochastic attention learning algorithm.
The final learning rule for the model is then the following:
\begin{multline}
 \frac{\partial L_s}{\partial W} \approx \frac{1}{N} \sum_{n=1}^{N} \bigg[ \frac{\partial \log p(\vy \mid \tilde{s}^n, \va)}{\partial W} +  \\
                               \lambda_r( \log p(\vy \mid \tilde{s}^n, \va) -  b) \frac{\partial \log p(\tilde{s}^n \mid \va)}{\partial W} + \lambda_e\frac{\partial H[\tilde{s}^n]}{\partial W} \bigg]
\nonumber 
\end{multline}
where, $\lambda_r$ and $\lambda_e$ are two hyper-parameters set by cross-validation. 
As pointed out and used in  \citet{Ba2014} and \citet{Mnih2014}, this is
formulation is equivalent to the REINFORCE learning rule~\citep{Williams92}, where the
reward for the attention choosing a sequence of actions is a real value proportional to the log likelihood
of the target sentence under the sampled attention trajectory. 

In making a hard choice at every point, $\phi\left( \left\{ \va_i \right\},
\left\{ \alpha_i \right\}\right)$ from Equation \ref{eq:context} is a function that
returns a sampled $\va_i$ at every point in time based upon a multinouilli
distribution parameterized by $\alpha$. 

\subsection{Deterministic ``Soft'' Attention}
\label{sec:det_attn}
Learning stochastic attention requires sampling the attention location $s_t$
each time, instead we can take the expectation of the context vector
$\hat{\vz}_t$ directly, 
\begin{align}
\label{eq:s_dist_soft}
    \mathbb{E}_{p(s_t|a)}[\hat{\vz}_t] =  \sum_{i=1}^L \alpha_{t,i} \va_{i}
\end{align} 
and formulate a deterministic attention model by computing a soft
attention weighted annotation vector $\phi\left( \left\{ \va_i \right\},
\left\{ \alpha_i \right\}\right) = \sum_i^L \alpha_i \va_i$ as introduced
by \citet{Bahdanau2014}. This corresponds to feeding in a
soft $\alpha$ weighted context into the system. The whole model is smooth and
differentiable under the deterministic attention, so learning end-to-end is
trivial by using standard back-propagation.

Learning the deterministic attention can also be understood as approximately
optimizing the marginal likelihood in Equation \ref{eq:mll} under the attention
location random variable $s_t$ from Sec.~\ref{sec:sto_attn}. The hidden activation of
LSTM $\vh_t$ is a linear projection of the stochastic context vector $\hat{\vz}_t$
followed by $\tanh$ non-linearity. To the first order Taylor approximation, the
expected value $\mathbb{E}_{p(s_t|a)}[\vh_t]$ is equal to computing $\vh_t$
using a single forward prop with the expected context vector
$\mathbb{E}_{p(s_t|a)}[\hat{\vz}_t]$. Considering Eq.~\ref{eq:p-out}, 
let $\vn_t = \vL_o(\vE\vy_{t-1} + \vL_h\vh_t+ \vL_z \hat{\vz}_t)$,
$\vn_{t,i}$ denotes $\vn_t$ computed by setting the random variable $\hat{\vz}$ value to $\va_i$.
We define the normalized weighted geometric mean for the softmax $k^{th}$ word
prediction:
\begin{align}
NWGM[p(y_t=k \mid \va)] &= {\prod_i \exp(n_{t,k,i})^{p(s_{t,i}=1|a)} \over \sum_j \prod_i \exp(n_{t,j,i})^{p(s_{t,i}=1|a)}} \nonumber \\
               &= {\exp(\mathbb{E}_{p(s_{t}|a)}[n_{t,k}]) \over \sum_j \exp(\mathbb{E}_{p(s_t|a)}[n_{t,j}])} \nonumber 
\end{align}
The equation above shows the normalized weighted geometric mean of the caption 
prediction can be approximated well by using the expected context vector, where 
$\mathbb{E}[\vn_t] = \vL_o(\vE\vy_{t-1} + \vL_h\mathbb{E}[\vh_t]+ \vL_z \mathbb{E}[\hat{\vz}_t])$.
It shows that the NWGM of a softmax unit is obtained by applying softmax to the
expectations of the underlying linear projections.   Also, from the results in
\cite{baldi2014dropout},  $NWGM[p(y_t=k \mid \va)] \approx \mathbb{E}[p(y_t=k \mid \va)]$ 
under softmax activation. That means the expectation of the outputs over all
possible attention locations induced by random variable $s_t$ is computed by
simple feedforward propagation with expected context vector
$\mathbb{E}[\hat{\vz}_t]$.  In other words, the deterministic attention model
is an approximation to the marginal likelihood over the attention locations.

\subsubsection{Doubly Stochastic Attention}
\label{section:ds_attn}

By construction, $\sum_i \alpha_{ti} = 1$ as they are the output of a softmax.
In training the deterministic version of our model we introduce a form of
doubly stochastic regularization, where we also encourage $\sum_t \alpha_{ti} \approx
1$. This can be interpreted as encouraging the model to pay equal attention to
every part of the image over the course of generation. In our experiments, we
observed that this penalty was important quantitatively to improving overall
BLEU score and that qualitatively this leads to more rich and descriptive
captions. In addition, the soft attention model predicts a gating scalar $\beta$ 
from previous hidden state $\vh_{t-1}$ at each time step $t$, such that, $\phi\left( \left\{ \va_i \right\}, \left\{ \alpha_i \right\}\right) = \beta \sum_i^L \alpha_i \va_i$, where $\beta_t = \sigma(f_{\beta}(\vh_{t-1}))$. We notice 
our attention weights put more emphasis on the objects in the images by 
including the scalar $\beta$.

\begin{table*}[!tph]
\label{table:results}
\caption{BLEU-{1,2,3,4}/METEOR metrics compared to other methods, $\dagger$ indicates a different split, (---) indicates an unknown metric, $\circ$ indicates the authors
kindly provided missing metrics by personal communication, $\Sigma$ indicates an ensemble, $a$ indicates using
AlexNet}

\vskip 0.15in
\centering
\begin{tabular}{cccccccc}
\cline{3-6}
& \multicolumn{1}{l|}{}      & \multicolumn{4}{c|}{\bf BLEU}                                                                                 &                             &  \\ \cline{1-7}
\multicolumn{1}{|c|}{Dataset} & \multicolumn{1}{c|}{Model} & \multicolumn{1}{c|}{BLEU-1} & \multicolumn{1}{c|}{BLEU-2} & \multicolumn{1}{c|}{BLEU-3} & \multicolumn{1}{c|}{BLEU-4} & \multicolumn{1}{c|}{METEOR} &  \\ \cline{1-7}
\multicolumn{1}{|c|}{Flickr8k} & \specialcell{ Google NIC\citep{Vinyals2014}$^\dagger$$^\Sigma$ \\ Log Bilinear \citep{Kiros2014a}$^\circ$ \\ Soft-Attention \\ Hard-Attention }
& \specialcell{63 \\ 65.6 \\ \bf 67 \\ \bf 67 }
& \specialcell{41 \\ 42.4 \\ 44.8 \\ \bf 45.7  }
& \specialcell{27 \\ 27.7 \\ 29.9 \\ \bf 31.4 }
& \specialcell{ ---  \\ 17.7 \\ 19.5 \\ \bf 21.3 }
& \multicolumn{1}{c|}{\specialcell{ --- \\ 17.31 \\ 18.93 \\ \bf 20.30 }} \ts\\
\cline{1-7}
\multicolumn{1}{|c|}{Flickr30k} & \specialcell{ Google NIC$^\dagger$$^\circ$$^\Sigma$  \\Log Bilinear\\ Soft-Attention \\ Hard-Attention }
& \specialcell{66.3  \\ 60.0 \\ 66.7 \\ \bf 66.9 }
& \specialcell{42.3  \\ 38 \\ 43.4 \\ \bf 43.9 }
& \specialcell{27.7  \\ 25.4 \\ 28.8 \\ \bf 29.6 }
& \specialcell{18.3 \\ 17.1 \\ 19.1 \\ \bf 19.9 }
& \multicolumn{1}{c|}{\specialcell{ --- \\ 16.88 \\ \bf 18.49 \\ 18.46}} \ts\\
\cline{1-7}
\multicolumn{1}{|c|}{COCO} & \specialcell{CMU/MS Research \citep{Chen2014}$^a$ \\ MS Research \citep{Fang2014}$^\dagger$$^a$ \\ BRNN \citep{Karpathy2014}$^\circ$ \\Google NIC$^\dagger$$^\circ$$^\Sigma$ \\ Log Bilinear$^\circ$ \\ Soft-Attention \\ Hard-Attention }
& \specialcell{--- \\ --- \\ 64.2 \\ 66.6 \\ 70.8 \\ 70.7 \\ \bf 71.8 }
& \specialcell{--- \\ --- \\ 45.1 \\ 46.1  \\ 48.9 \\ 49.2 \\ \bf 50.4  }
& \specialcell{--- \\ --- \\ 30.4 \\ 32.9 \\ 34.4 \\ 34.4 \\ \bf 35.7 }
& \specialcell{--- \\ --- \\ 20.3 \\ 24.6  \\ 24.3 \\ 24.3  \\ \bf 25.0 }
& \multicolumn{1}{c|}{\specialcell{ 20.41 \\ 20.71 \\ --- \\ --- \\ 20.03 \\ \bf 23.90 \\ 23.04 }} \ts\\
\cline{1-7}
\end{tabular}
\vskip -0.1in
\end{table*}

Concretely, the model is trained end-to-end by minimizing the following penalized negative
log-likelihood:
\begin{gather}
L_d =  -\log(P(\textbf{y}|\textbf{x})) + \lambda \sum_i^{L}(1 - \sum_t^C \alpha_{ti})^2
\end{gather}
\subsection{Training Procedure}

Both variants of our attention model were trained with stochastic gradient
descent using adaptive learning rate algorithms. For the Flickr8k dataset, we
found that RMSProp \citep{Tieleman2012} worked best, while for Flickr30k/MS
COCO dataset we used the recently proposed Adam algorithm~\citep{Kingma2014} . 

To create the annotations $a_i$ used by our decoder, we used the Oxford VGGnet
\citep{Simonyan14} pre-trained on ImageNet without finetuning. In principle however,
any encoding function could be used. In addition, with enough data, we could also train
the encoder from scratch (or fine-tune) with the rest of the model.
In our experiments we use the 14$\times$14$\times$512 feature map
of the fourth convolutional layer before max pooling. This means our
decoder operates on the flattened 196 $\times$ 512 (i.e $L\times D$) encoding.
 
As our implementation requires time proportional to the length of the longest
sentence per update, we found training on a random group of captions to be
computationally wasteful. To mitigate this problem, in preprocessing we build a
dictionary mapping the length of a sentence to the corresponding subset of
captions. Then, during training we randomly sample a length and retrieve a
mini-batch of size 64 of that  length. We found that this greatly improved
convergence speed with no noticeable diminishment in performance. On our
largest dataset (MS COCO), our soft attention model took less than 3 days to
train on an NVIDIA Titan Black GPU. 

In addition to dropout~\citep{Srivastava14}, the only other regularization
strategy we used was early stopping on BLEU score. We observed a breakdown in
correlation between the validation set log-likelihood and BLEU in the later stages
of training during our experiments. Since BLEU is the most commonly reported
metric, we used BLEU on our validation set for model selection.

In our experiments with
soft attention, we also used Whetlab\footnote{\url{https://www.whetlab.com/}} \citep{Snoek2012,Snoek2014} in our
Flickr8k experiments. Some of the intuitions we gained from hyperparameter
regions it explored were especially important in our Flickr30k and COCO experiments. 

We make our code for these models based in Theano \citep{Bergstra2010}
publicly available upon publication to encourage future research in this
area. 

\section{Experiments}
We describe our experimental methodology and quantitative results which validate the effectiveness
of our model for caption generation. 

\subsection{Data}
We report results on the popular Flickr8k and Flickr30k dataset which has 8,000
and 30,000 images respectively as well as the more challenging Microsoft COCO
dataset which has 82,783 images. The Flickr8k/Flickr30k dataset both come with
5 reference sentences per image, but for the MS COCO dataset, some of the
images have references in excess of 5 which for consistency across
our datasets we discard. We applied only basic tokenization to MS COCO so that it is
consistent with the tokenization present in Flickr8k and Flickr30k. For all our
experiments, we used a fixed vocabulary size of 10,000.

Results for our attention-based architecture are reported in Table
\ref{table:results}. We report results with the frequently used BLEU
metric\footnote{We verified that our BLEU evaluation code matches the authors of
\citet{Vinyals2014}, \citet{Karpathy2014} and \citet{Kiros2014b}. For fairness, we only compare against
results for which we have verified that our BLEU evaluation code is the same. With
the upcoming release of the COCO evaluation server, we will include comparison
results with all other recent image captioning models.} which is the standard in the
caption generation literature. We report BLEU from 1 to 4 without a brevity
penalty. There has been, however, criticism of BLEU, so in addition we report
another common metric METEOR \citep{Denkowski2014}, and compare whenever
possible. 

\subsection{Evaluation Procedures}

A few challenges exist for comparison, which we explain here. The first is a
difference in choice of convolutional feature extractor. For identical decoder
architectures, using more recent architectures such as GoogLeNet or Oxford VGG
\citet{Szegedy2014}, \citet{Simonyan14} can give a boost in performance over
using the AlexNet \citep{Krizhevsky2012}. In our evaluation, we compare directly
only with results which use the comparable GoogLeNet/Oxford VGG features, but
for METEOR comparison we note some results that use AlexNet.

The second challenge is a single model versus ensemble comparison. While
other methods have reported performance boosts by using ensembling, in our
results we report a single model performance.

Finally, there is challenge due to differences between dataset splits. In our
reported results, we use the pre-defined splits of Flickr8k. However, one
challenge for the Flickr30k and COCO datasets is the lack of standardized
splits. As a result, we report with the publicly available
splits\footnote{\url{http://cs.stanford.edu/people/karpathy/deepimagesent/}}
used in previous work \citep{Karpathy2014}. In our experience, differences in
splits do not make a substantial difference in overall performance, but
we note the differences where they exist.

\subsection{Quantitative Analysis}
\label{section:results}
In Table \ref{table:results}, we provide a summary of the experiment validating
the quantitative effectiveness of attention. We obtain state of the art
performance on the Flickr8k, Flickr30k and MS COCO. In addition, we note
that in our experiments we are able to significantly improve the state of the
art performance METEOR on MS COCO that we speculate is connected to
some of the regularization techniques we used \ref{section:ds_attn} and our
lower level representation. Finally, we also note that we are
able to obtain this performance using a single model without an ensemble.

\subsection{Qualitative Analysis: Learning to attend}
\label{section:viz}
By visualizing the attention component learned by the model, we are able to add
an extra layer of interpretability to the output of the model
(see Fig. 1).  Other systems that have done this rely on object
detection systems to produce candidate alignment targets~\citep{Karpathy2014}.
Our approach is much more flexible, since the model can attend to ``non
object'' salient regions. 

The 19-layer OxfordNet uses stacks of 3x3 filters meaning the only time the
feature maps decrease in size are due to the max pooling layers. The input
image is resized so that the shortest side is 256 dimensional with preserved
aspect ratio. The input to the convolutional network is the center cropped
224x224 image. Consequently, with 4 max pooling layers we get an output
dimension of the top convolutional layer of 14x14. Thus in order to visualize
the attention weights for the soft model, we simply upsample the weights by a
factor of $2^4$ = 16 and apply a Gaussian filter. We note that the receptive
fields of each of the 14x14 units are highly overlapping. 

As we can see in Figure 2 and 3, the model learns alignments
that correspond very strongly with human intuition. Especially in the examples of
mistakes, we see that it is possible to exploit such visualizations to get an
intuition as to why those mistakes were made. We provide a more extensive list of
visualizations in Appendix \ref{section:appendix} for the reader.

\section{Conclusion}

We propose an attention based approach that gives state of the art performance
on three benchmark datasets using the BLEU and METEOR metric. We also show how the
learned attention can be exploited to give more interpretability into the
models generation process, and demonstrate that the learned alignments
correspond very well to human intuition. We hope that the results of this paper
will encourage future work in using visual attention. We also expect that the
modularity of the encoder-decoder approach combined with attention to have
useful applications in other domains. 

\section*{Acknowledgments}
\label{sec:ack}

The authors would like to thank the developers of Theano~\citep{Bergstra2010,
Bastien2012}.  We acknowledge the support of the following
organizations for research funding and computing support: the Nuance Foundation, NSERC, Samsung,
Calcul Qu\'{e}bec, Compute Canada, the Canada Research Chairs and CIFAR. The authors
would also like to thank Nitish Srivastava for assistance with his ConvNet
package as well as preparing the Oxford convolutional network and Relu Patrascu
for helping with numerous infrastructure related problems.

%% file: appendix.tex
\appendix
\normalsize
\onecolumn
\section{Appendix}
\label{section:appendix}
Visualizations from our ``hard'' (a) and ``soft'' (b) attention
model. \textit{White} indicates the regions where the model roughly attends to (see section \ref{section:viz}).
\vskip -0.14in
\begin{figure*}[!ht]
\begin{center}
\centerline{\includegraphics[width=6.75in]{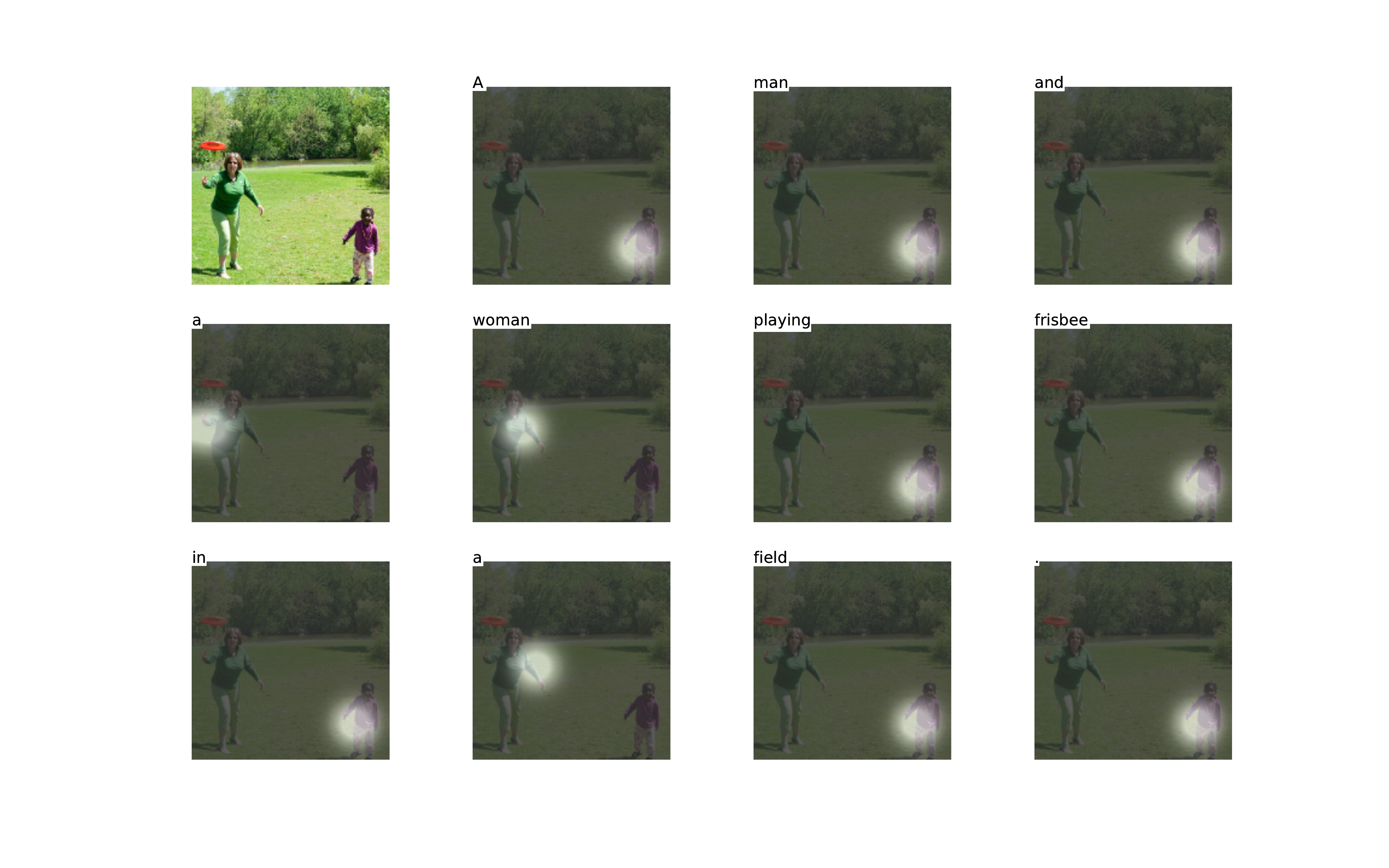}}
\vskip -0.3in
(a) A man and a woman playing frisbee in a field.
\centerline{\includegraphics[width=6.75in]{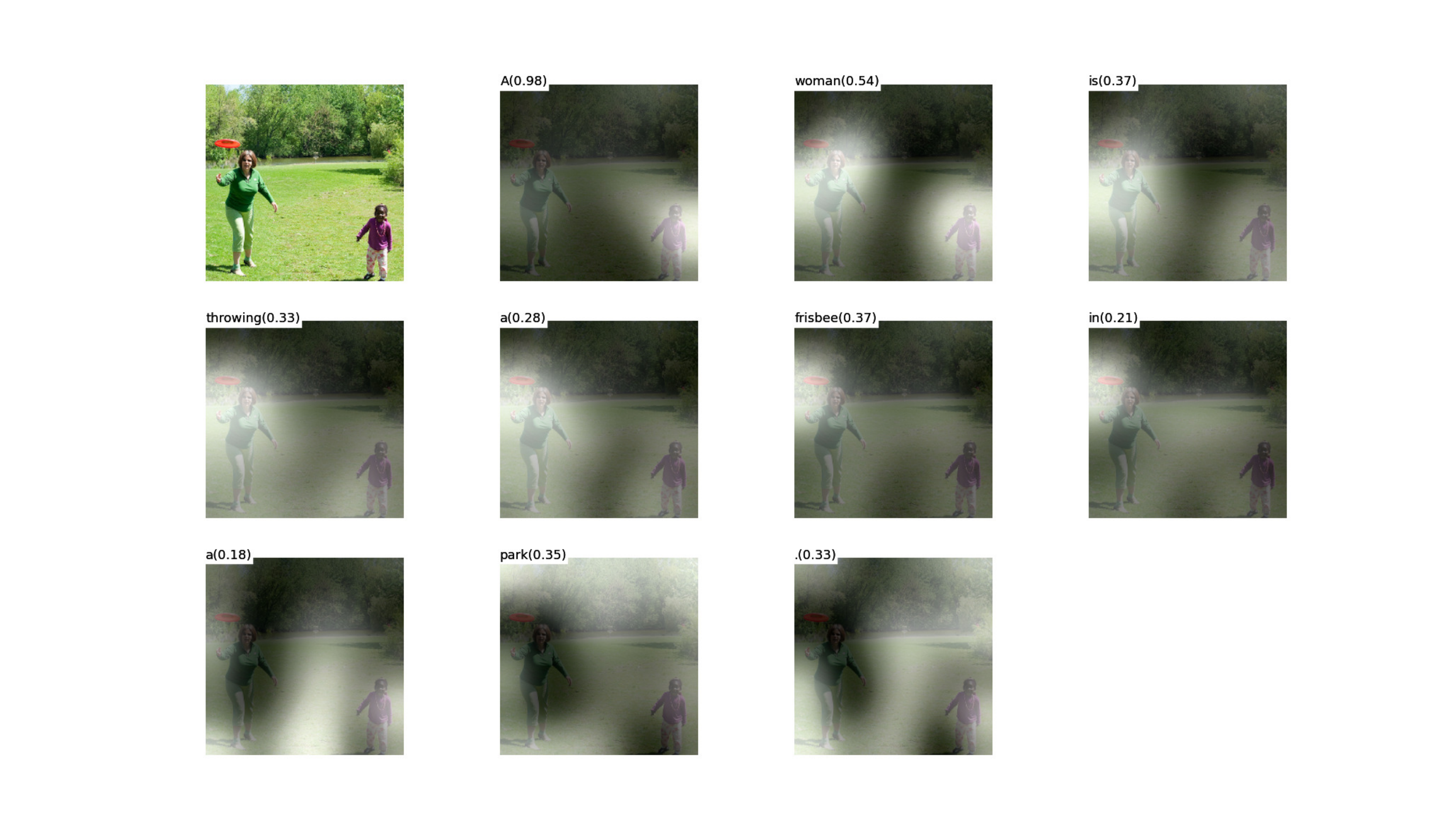}}
\vskip -0.2in
(b) A woman is throwing a frisbee in a park.
\caption{}
\label{figure:im61}
\vskip -3in
\end{center}
\end{figure*} 

\begin{figure*}[ht]
\vskip 0.2in
\begin{center}
\centerline{\includegraphics[width=7.3in]{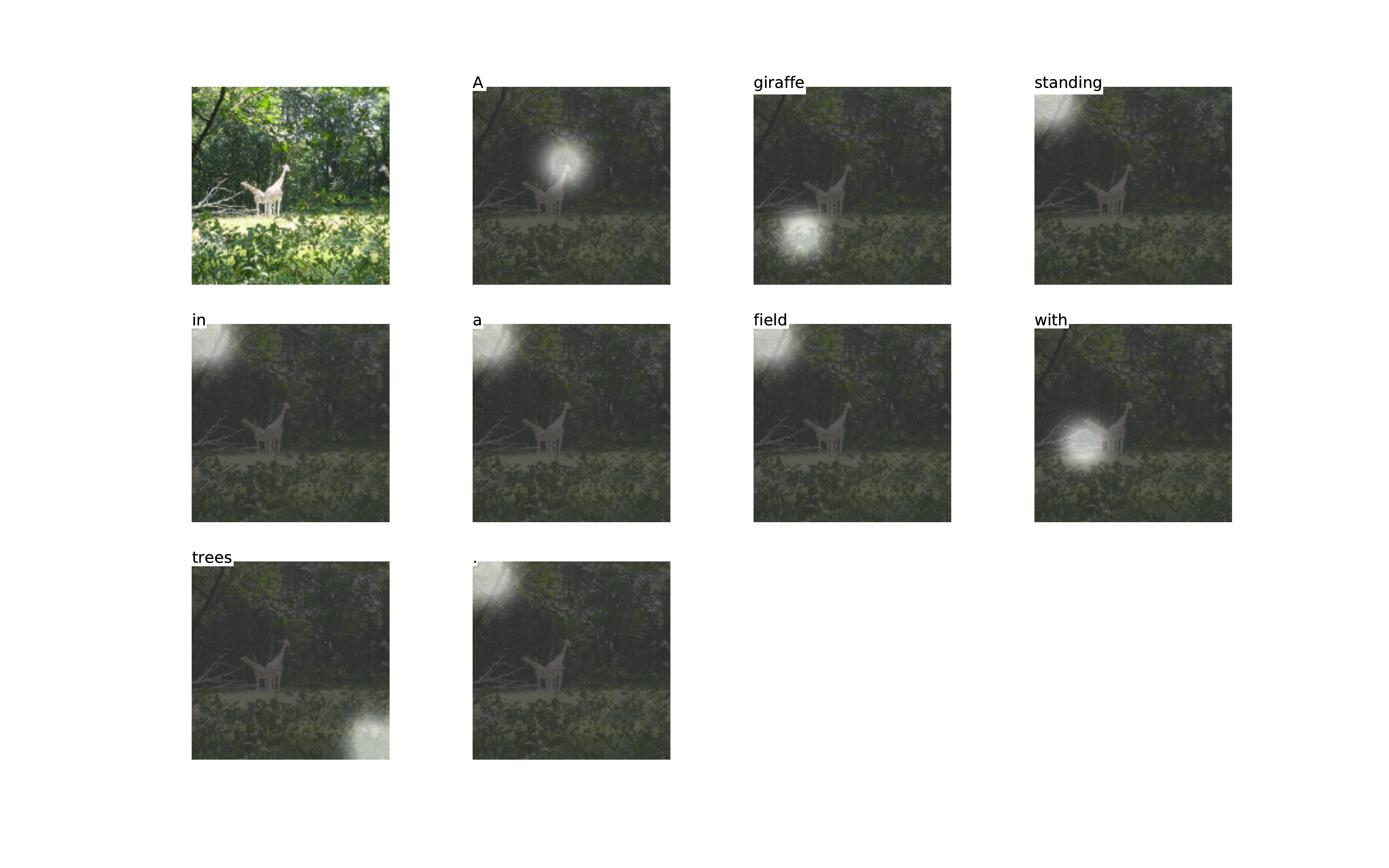}}
\vskip -0.2in
(a) A giraffe standing in the field with trees.
\centerline{\includegraphics[width=7.3in]{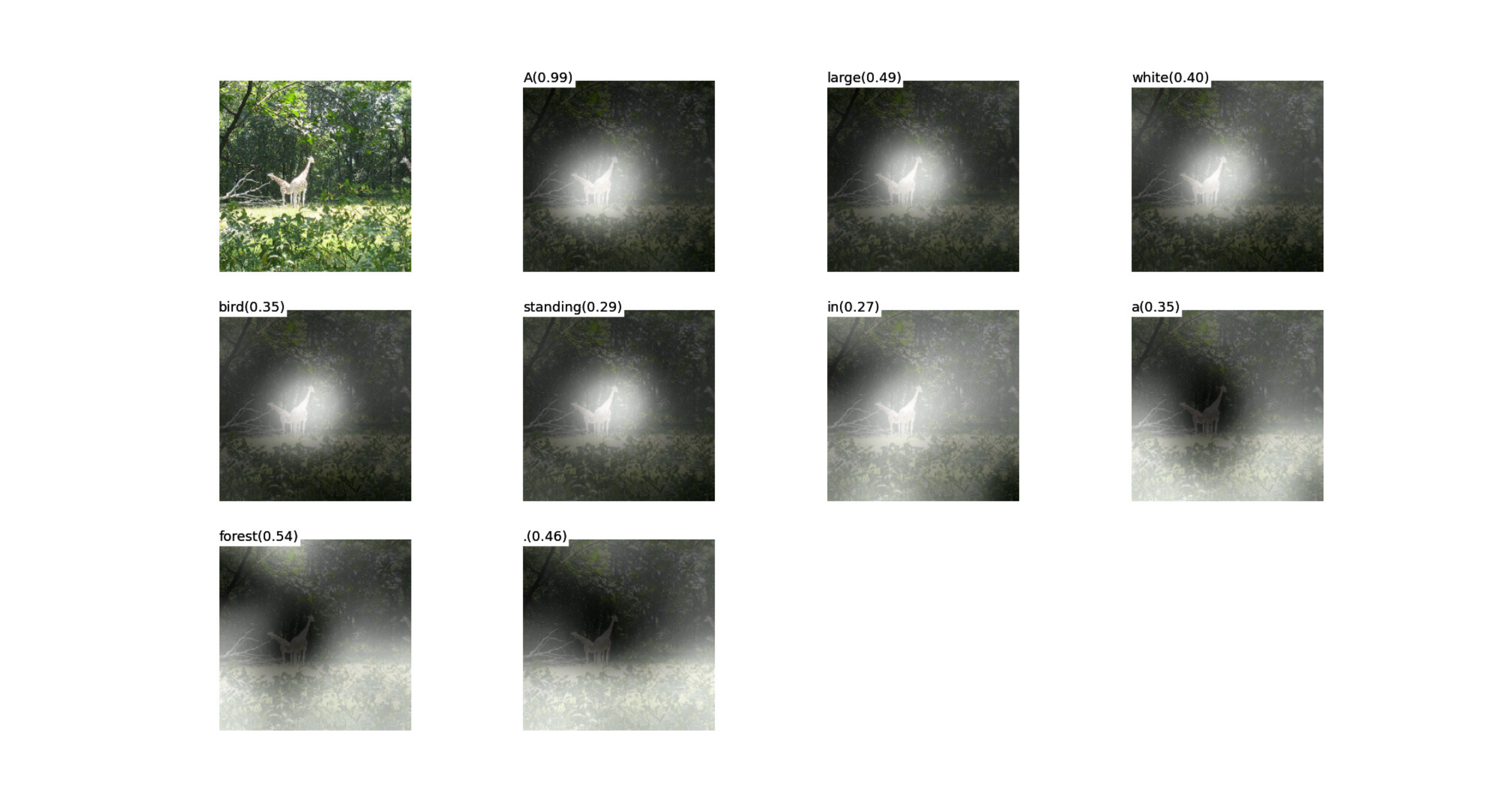}}
(b) A large white bird standing in a forest.
\caption{}
\label{figure:im1038}
\end{center}
\vskip -0.4in
\end{figure*} 

\begin{figure*}[ht]
\begin{center}
\centerline{\includegraphics[width=7in]{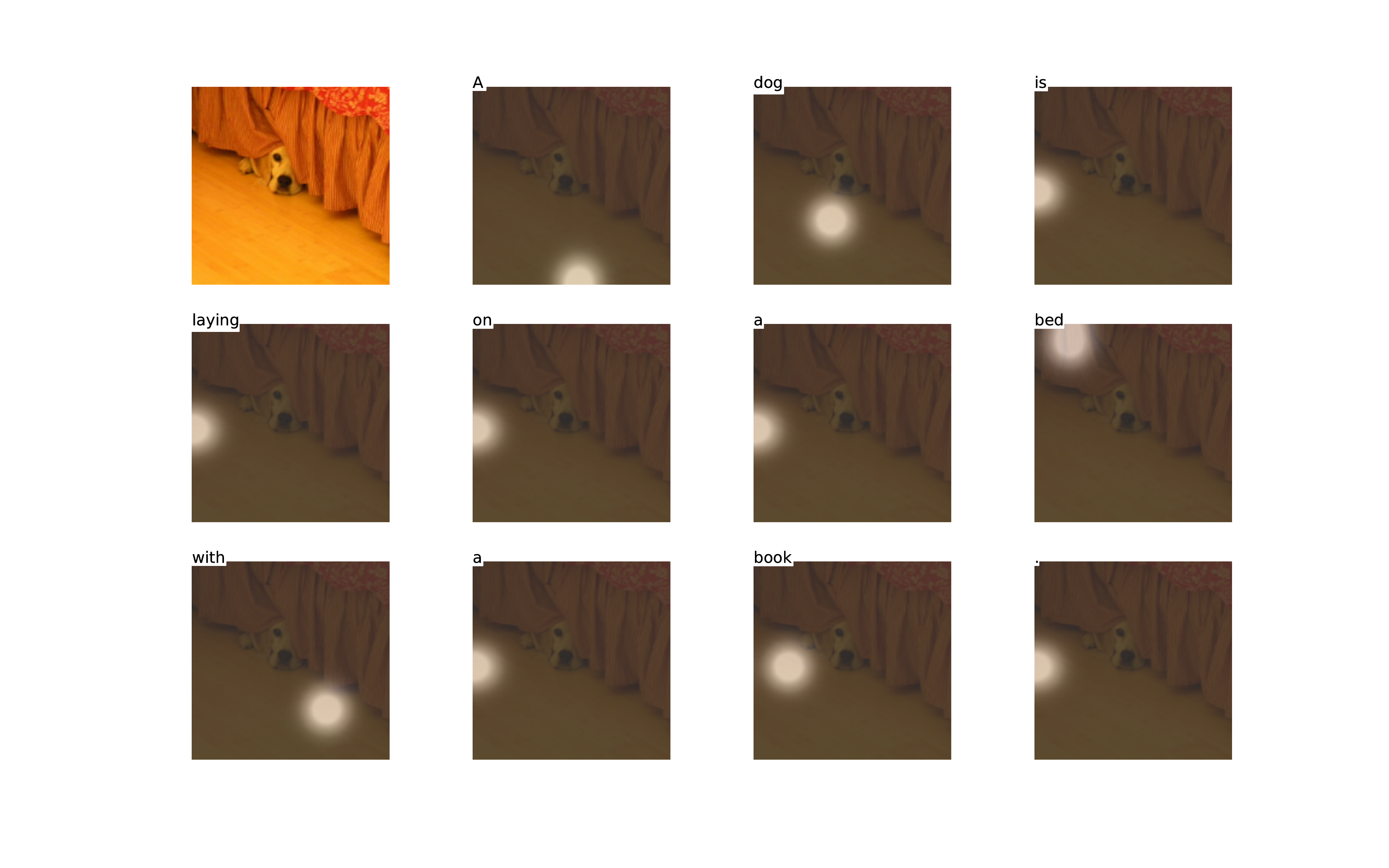}}
\vskip -0.2in
(a) A dog is laying on a bed with a book.
\centerline{\includegraphics[width=7in]{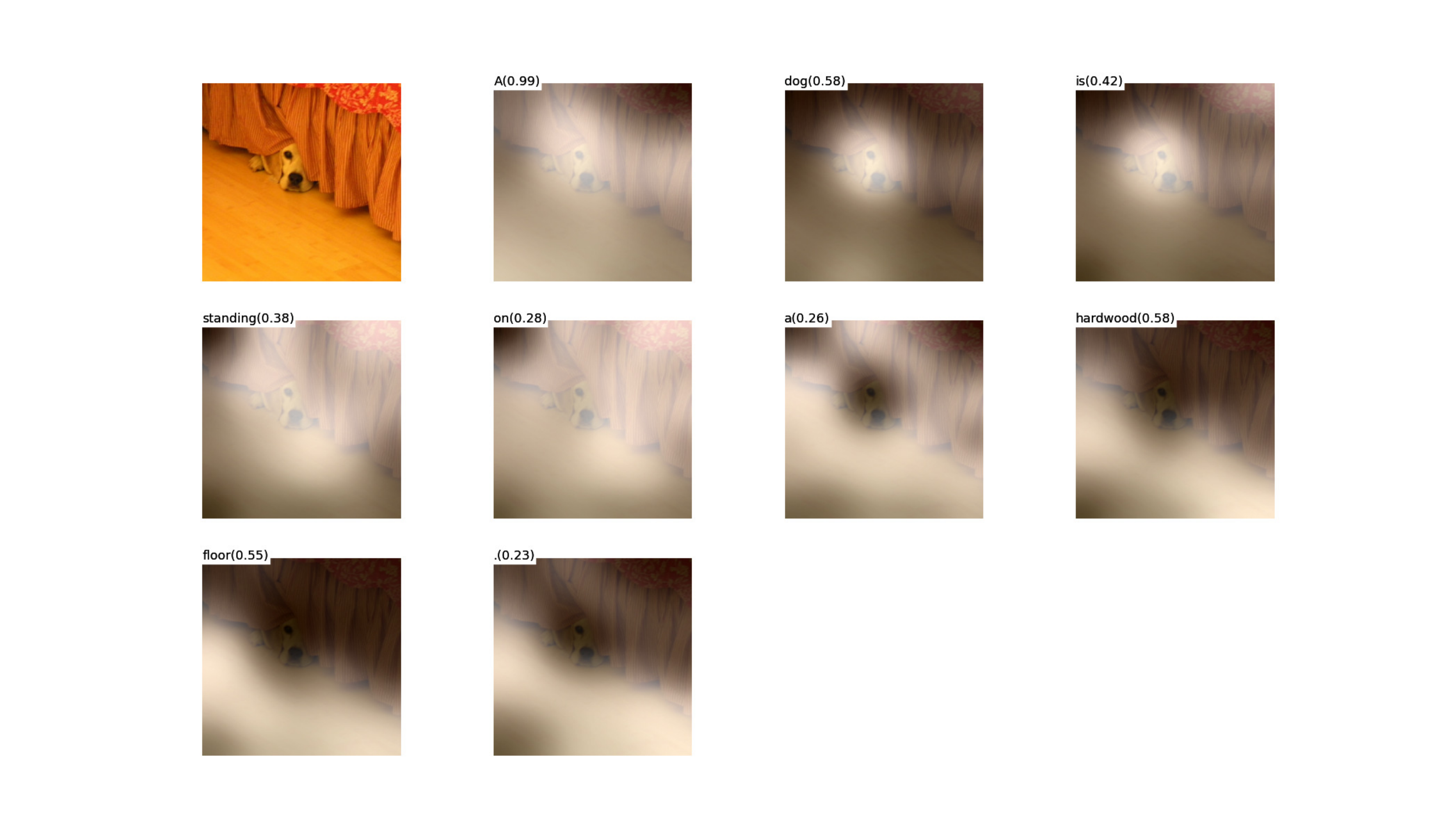}}
(b) A dog is standing on a hardwood floor.
\caption{}
\label{figure:im570}
\end{center}
\vskip -0.4in
\end{figure*} 

\begin{figure*}[ht]
\begin{center}
\centerline{\includegraphics[width=7in]{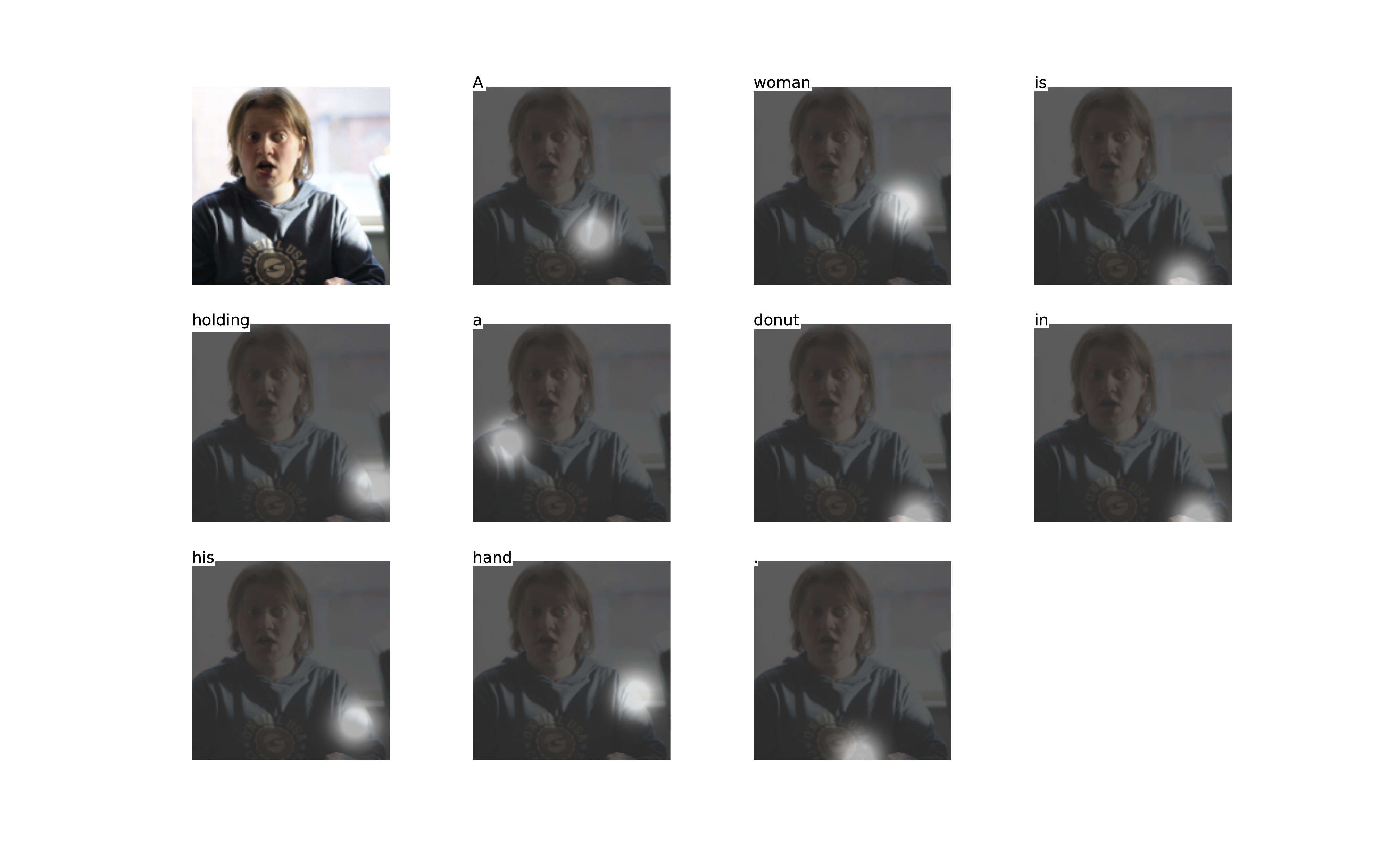}}
\vskip -0.3in
(a) A woman is holding a donut in his hand.
\centerline{\includegraphics[width=7in]{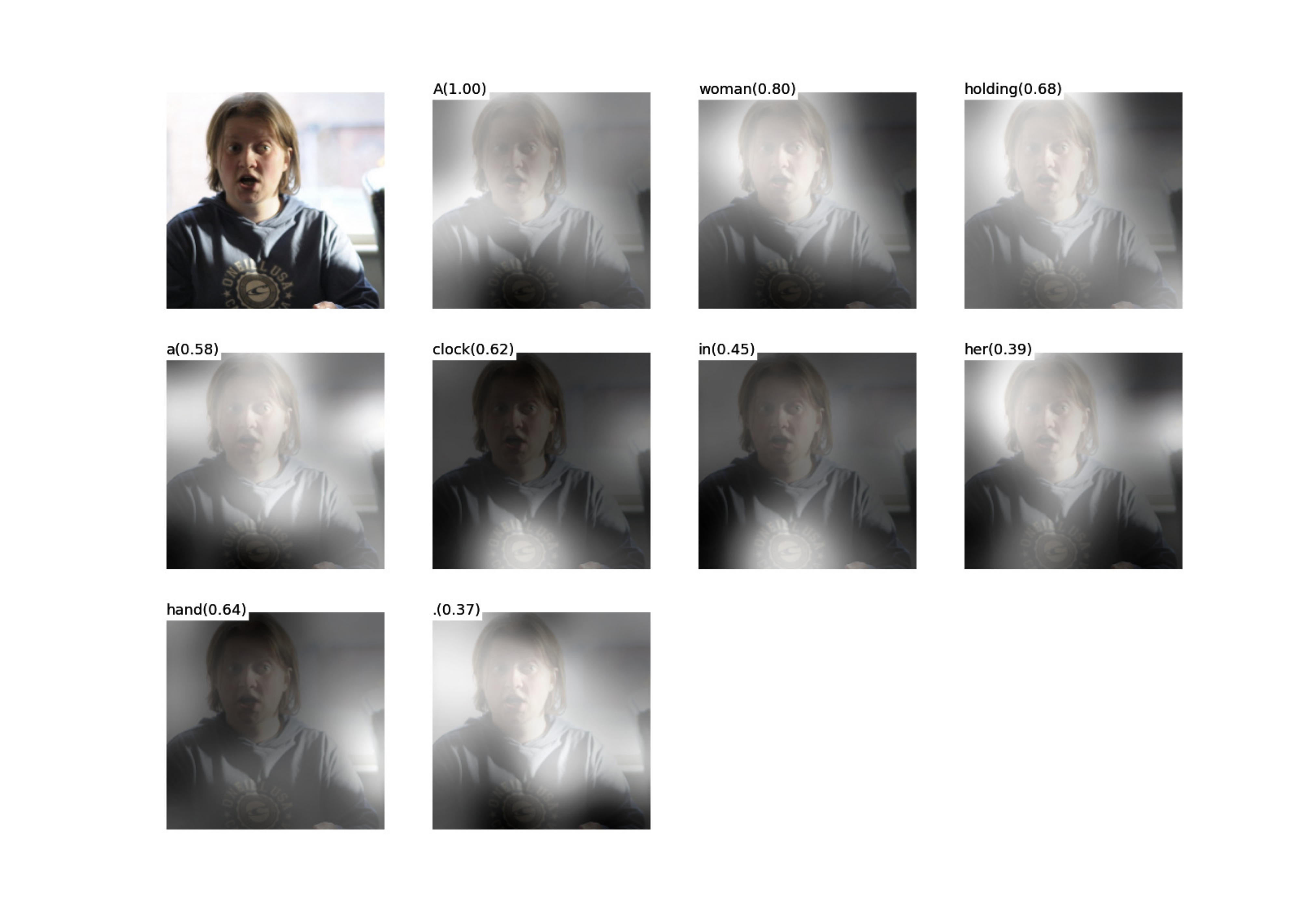}}
\vskip -0.4in
(b) A woman holding a clock in her hand.
\caption{}
\label{figure:im352}
\end{center}
\vskip -0.4in
\end{figure*}

\begin{figure*}[ht]
\begin{center}
\vskip -0.1in
\centerline{\includegraphics[width=8in]{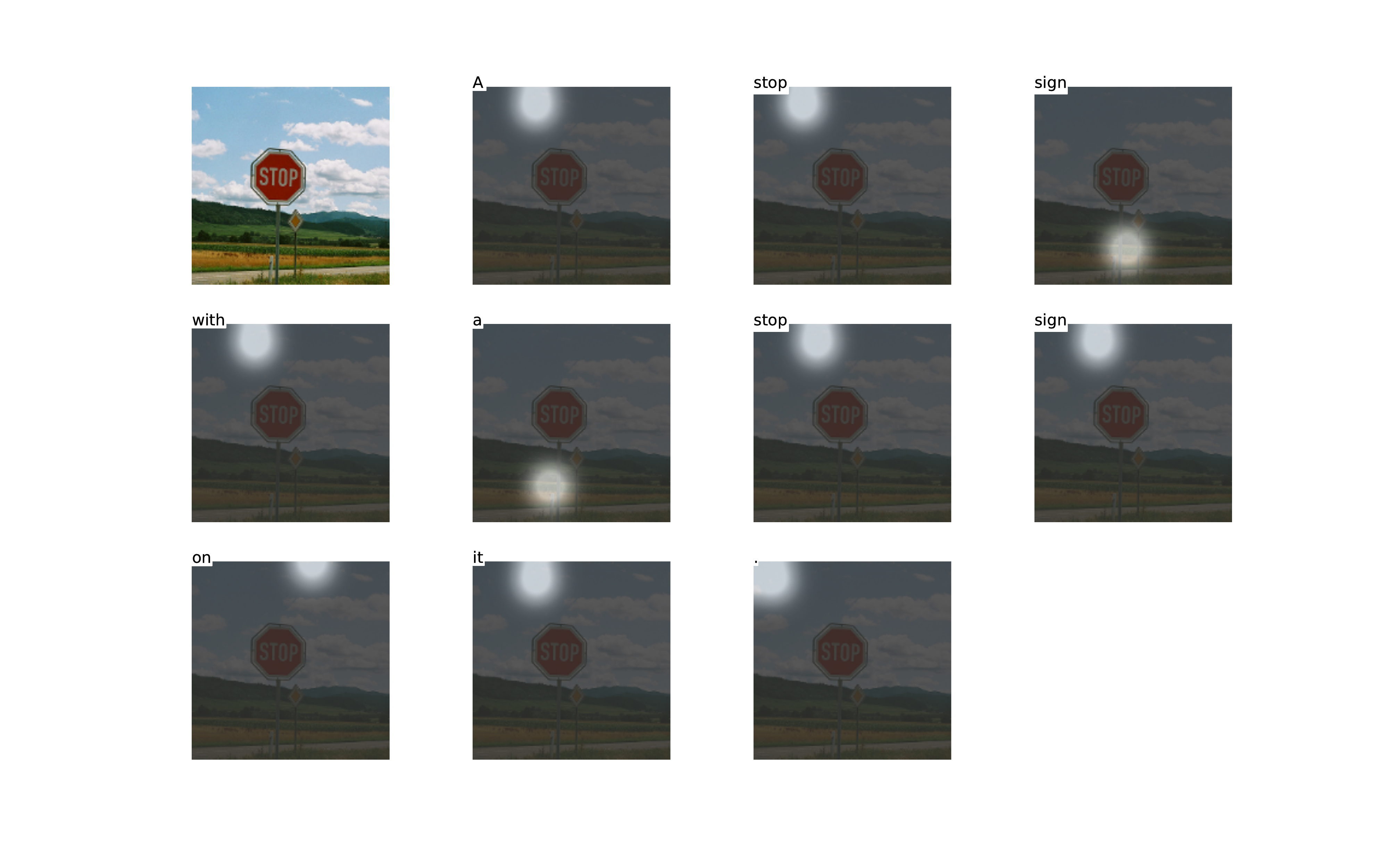}}
\vskip -0.5in
(a) A stop sign with a stop sign on it.
\centerline{\includegraphics[width=9in]{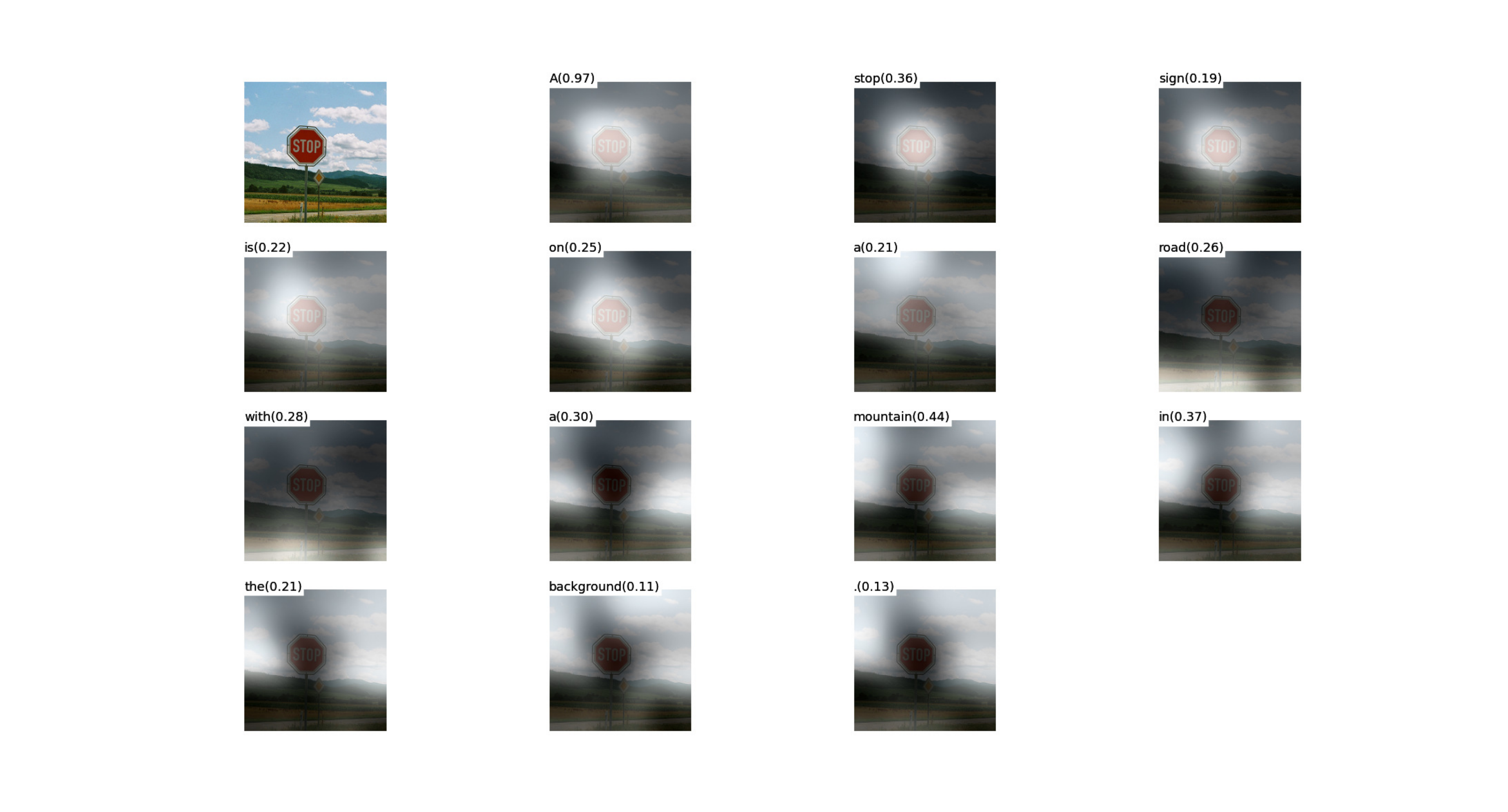}}
\vskip -0.2in
(b) A stop sign is on a road with a mountain in the background.
\caption{}
\label{figure:im1304}
\end{center}
\vskip -0.4in
\end{figure*}

\begin{figure*}[ht]
\begin{center}
\centerline{\includegraphics[width=7.5in]{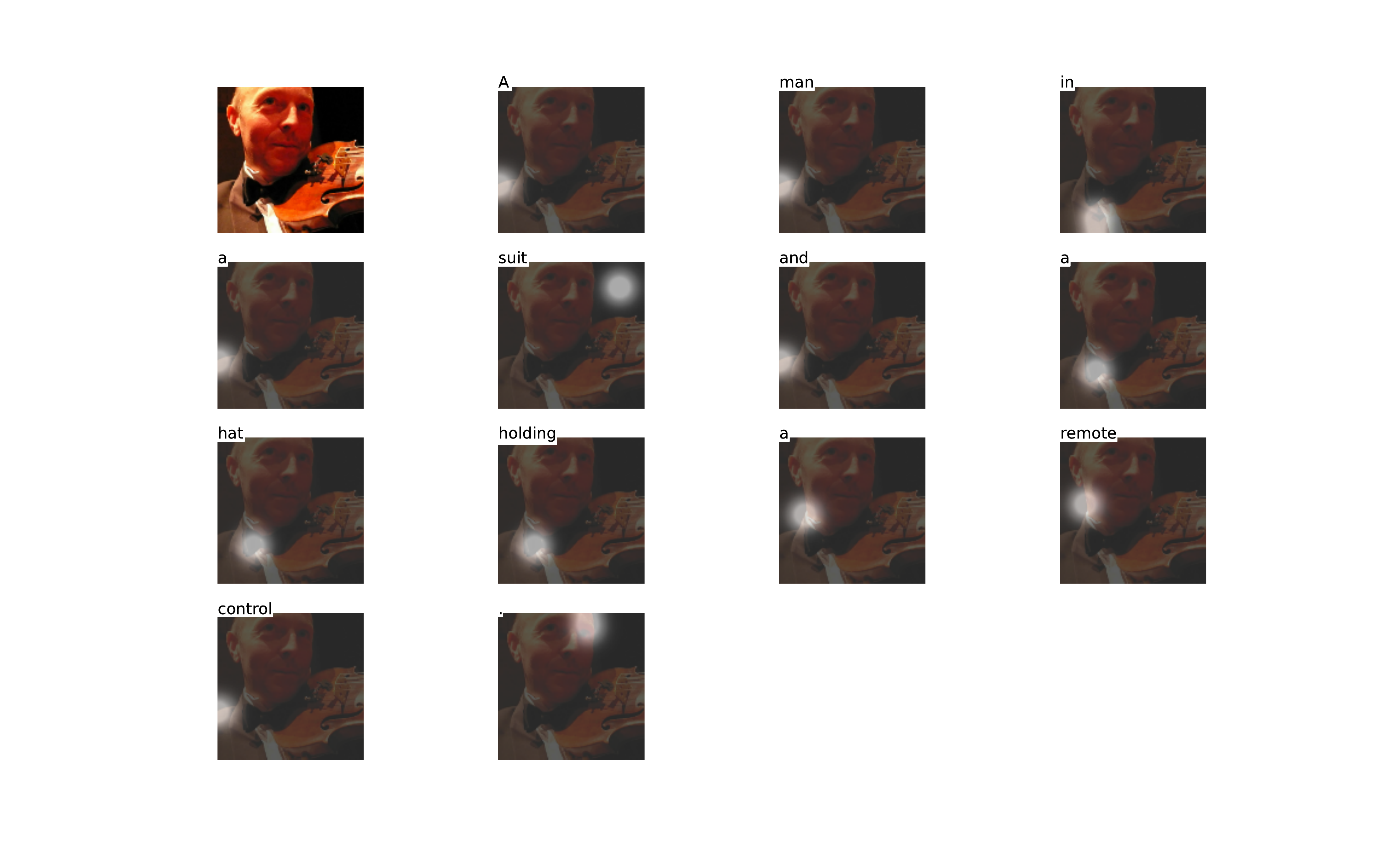}}
\vskip -0.5in
(a) A man in a suit and a hat holding a remote control.
\centerline{\includegraphics[width=7.5in]{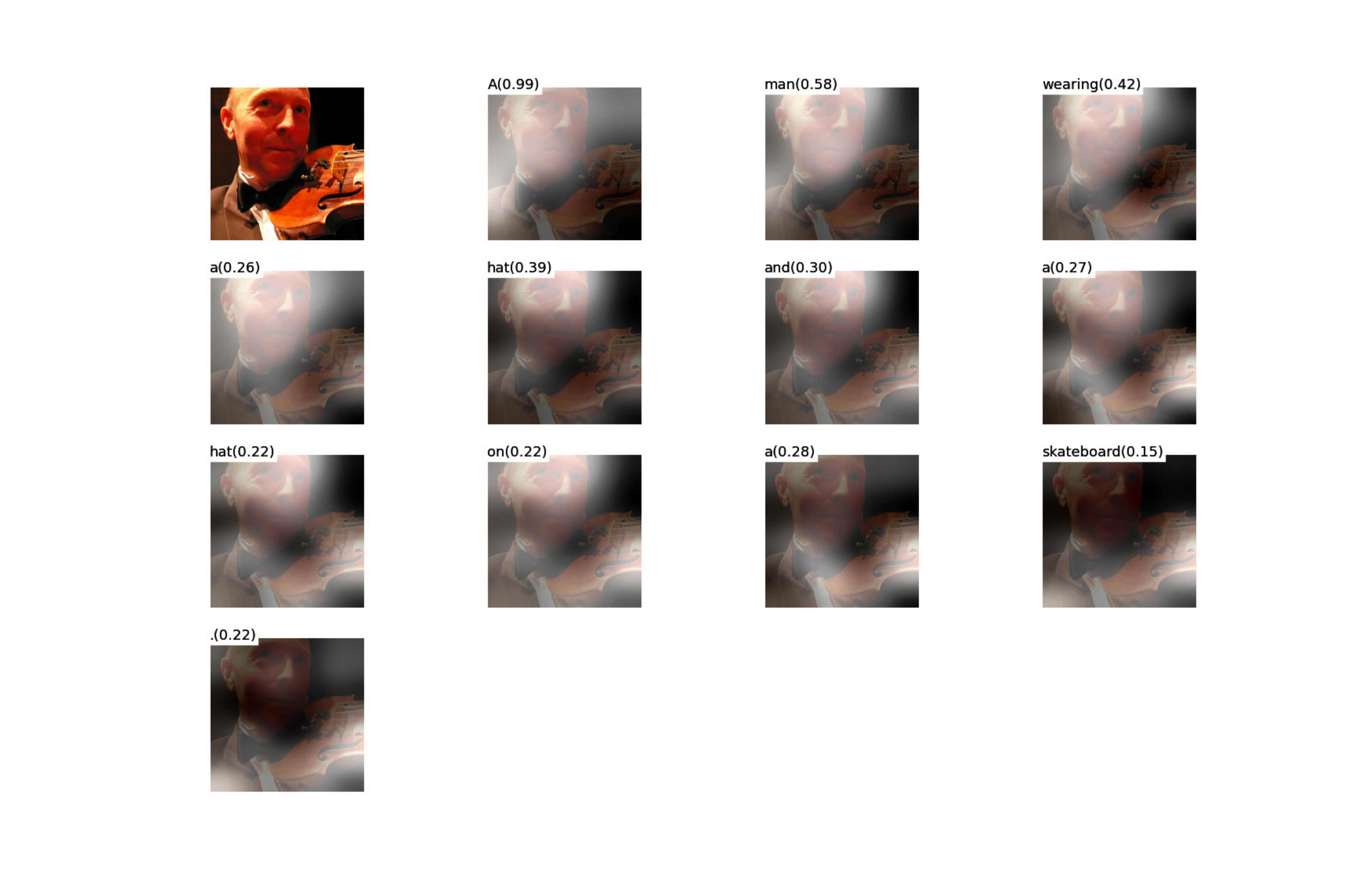}}
\vskip -0.2in
(b) A man wearing a hat and a hat on a skateboard.
\caption{}
\label{figure:im1066}
\end{center}
\vskip -0.7in
\end{figure*}

\begin{figure*}[ht]
\vskip 0.2in
\begin{center}
\centerline{\includegraphics[width=7.5in]{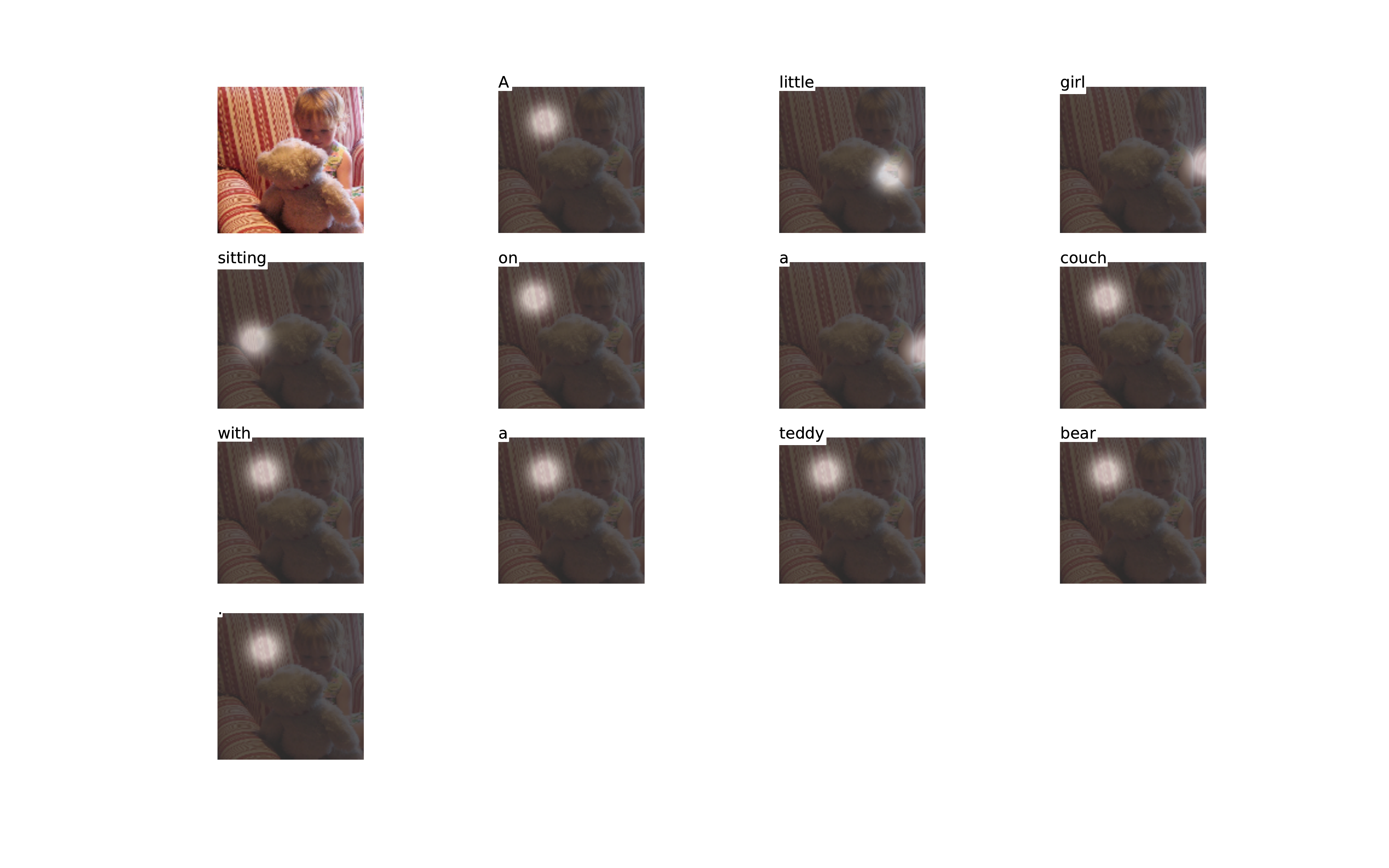}}
\vskip -0.4in
(a) A little girl sitting on a couch with a teddy bear.
\centerline{\includegraphics[width=7.5in]{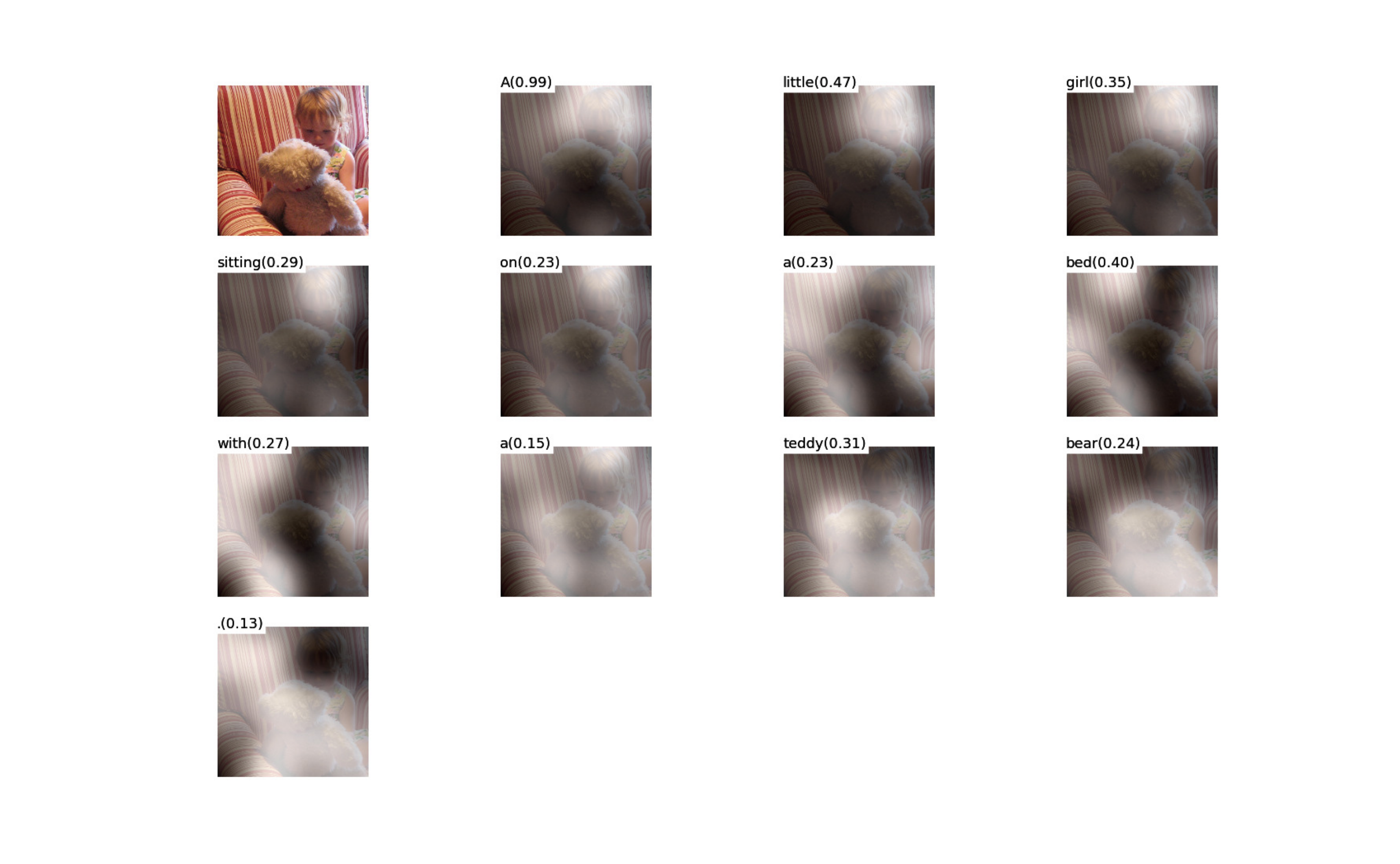}}
\vskip -0.2in
(b) A little girl sitting on a bed with a teddy bear.
\label{figure:im120}
\end{center}
\vskip -0.4in
\end{figure*}

\begin{figure*}[ht]
\begin{center}
\centerline{\includegraphics[width=7in]{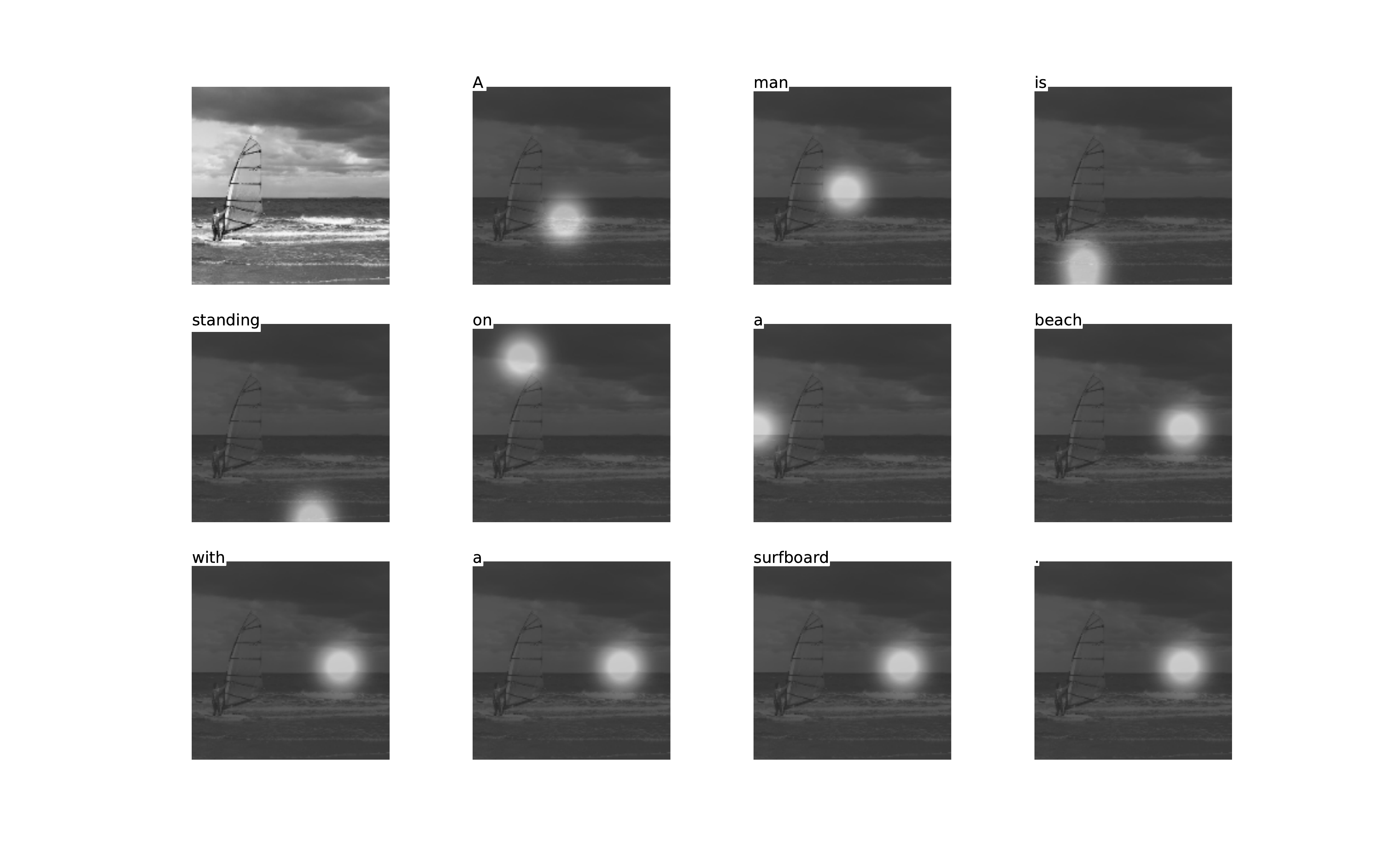}}
\vskip -0.4in
(a) A man is standing on a beach with a surfboard.
\centerline{\includegraphics[width=7in]{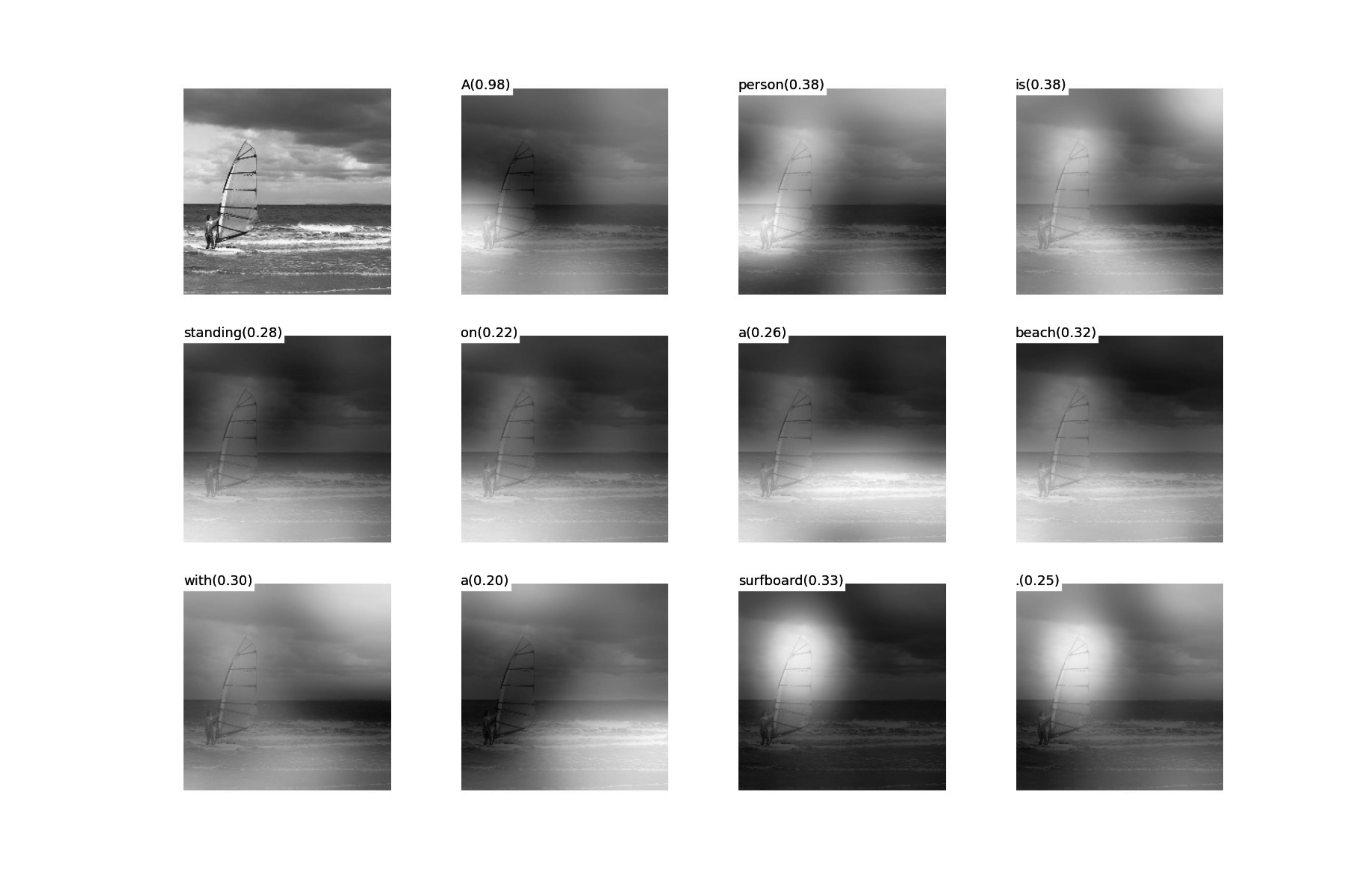}}
\vskip -0.2in
(b) A person is standing on a beach with a surfboard.
\label{figure:im861_3884}
\end{center}
\vskip -0.8in
\end{figure*}

\begin{figure*}[ht]
\begin{center}
\centerline{\includegraphics[width=7.5in]{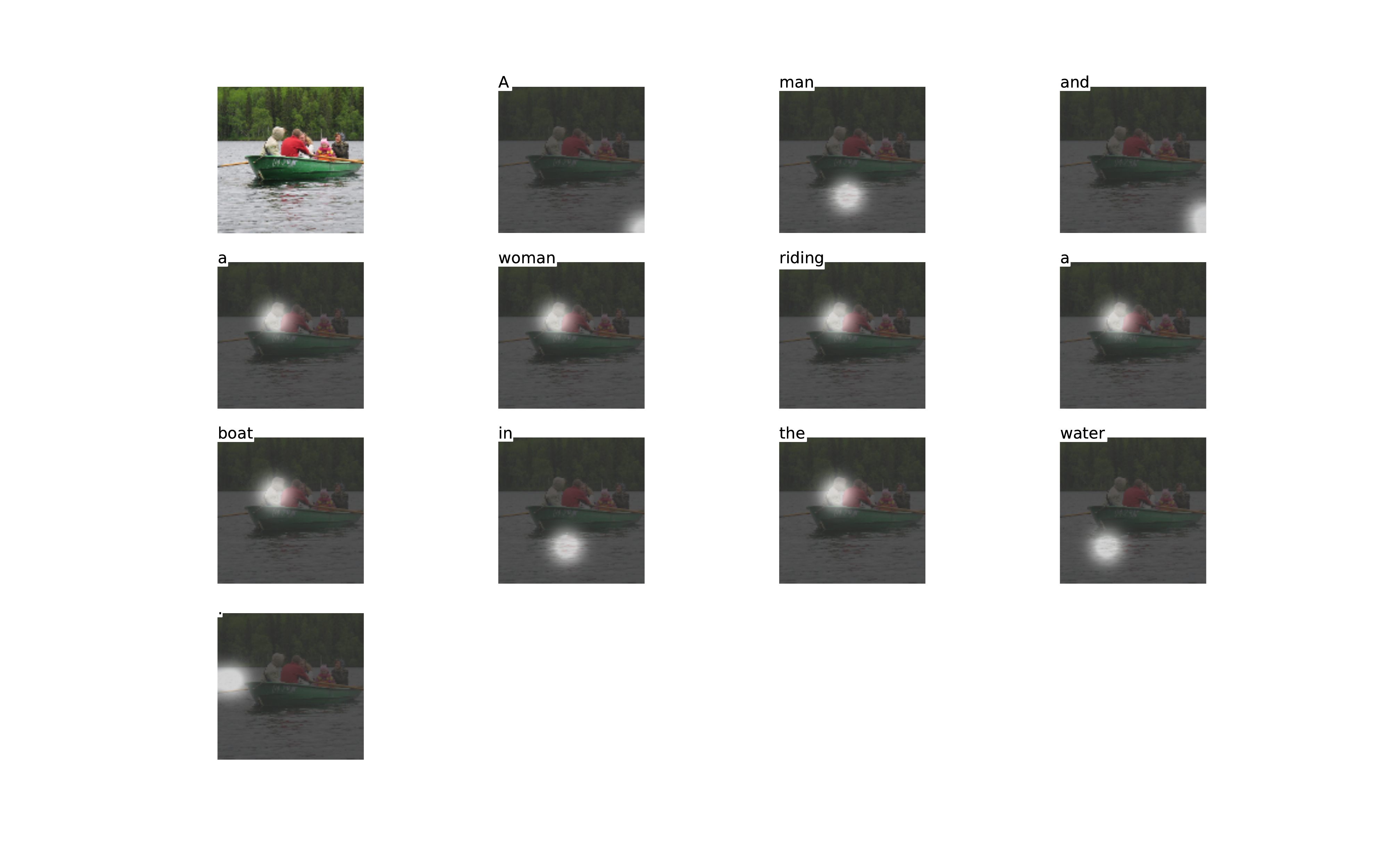}}
\vskip -0.4in
(a) A man and a woman riding a boat in the water.
\centerline{\includegraphics[width=7.5in]{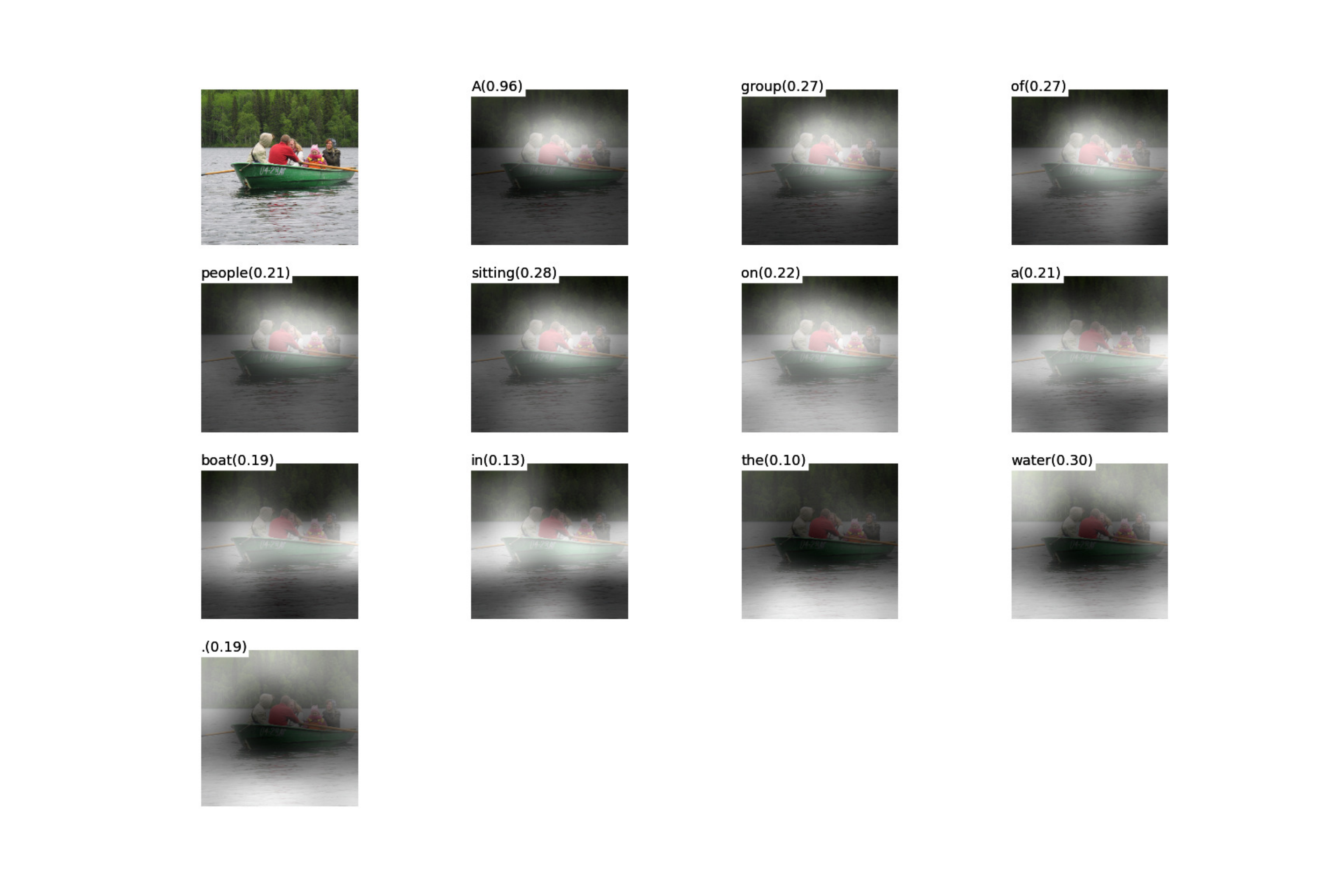}}
\vskip -0.4in
(b) A group of people sitting on a boat in the water.
\caption{}
\label{figure:im1322_1463_3861_4468}
\end{center}
\vskip -0.4in
\end{figure*}

\begin{figure*}[ht]
\begin{center}
\centerline{\includegraphics[width=8in]{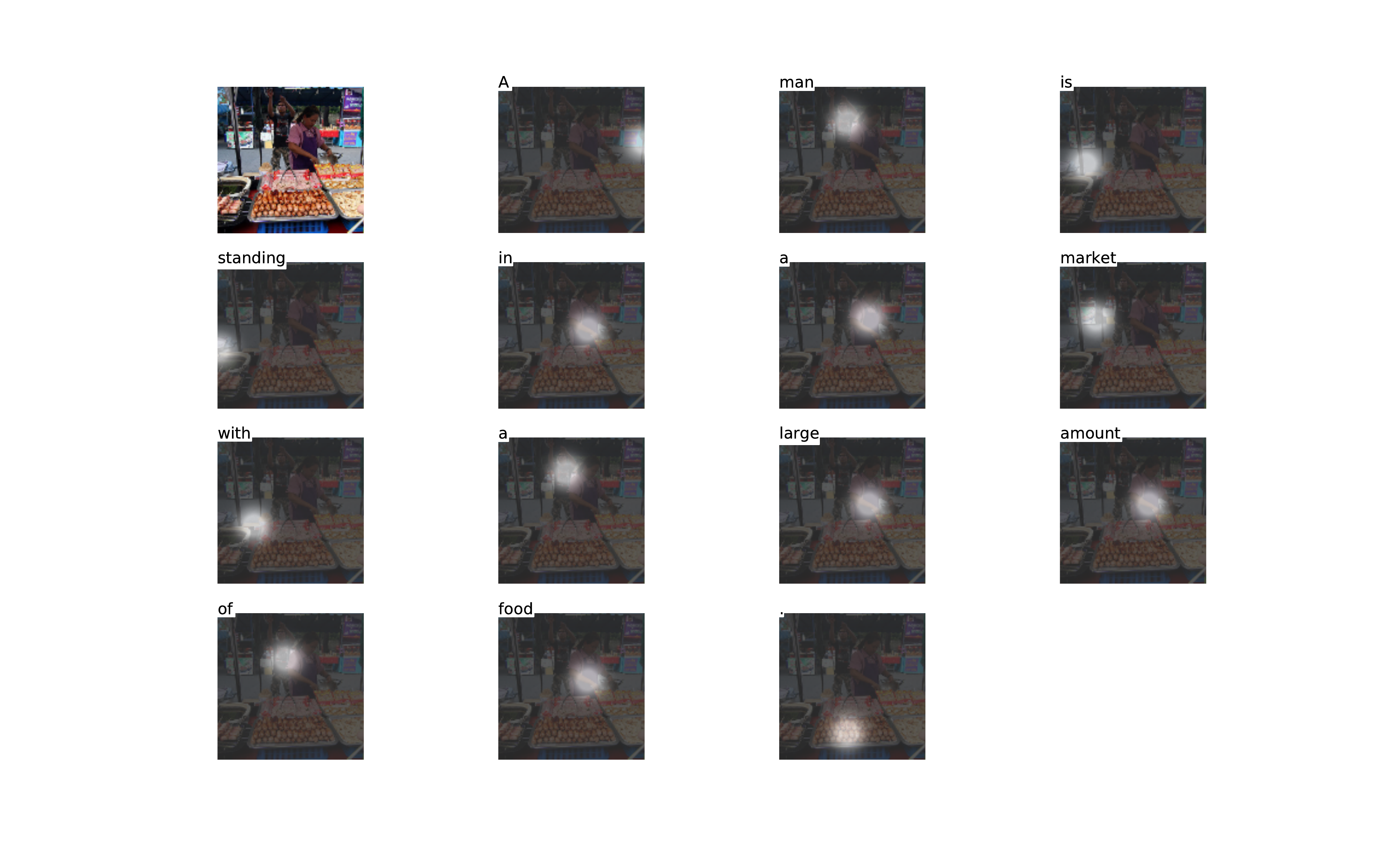}}
\vskip -0.5in
(a) A man is standing in a market with a large amount of food.
\centerline{\includegraphics[width=8in]{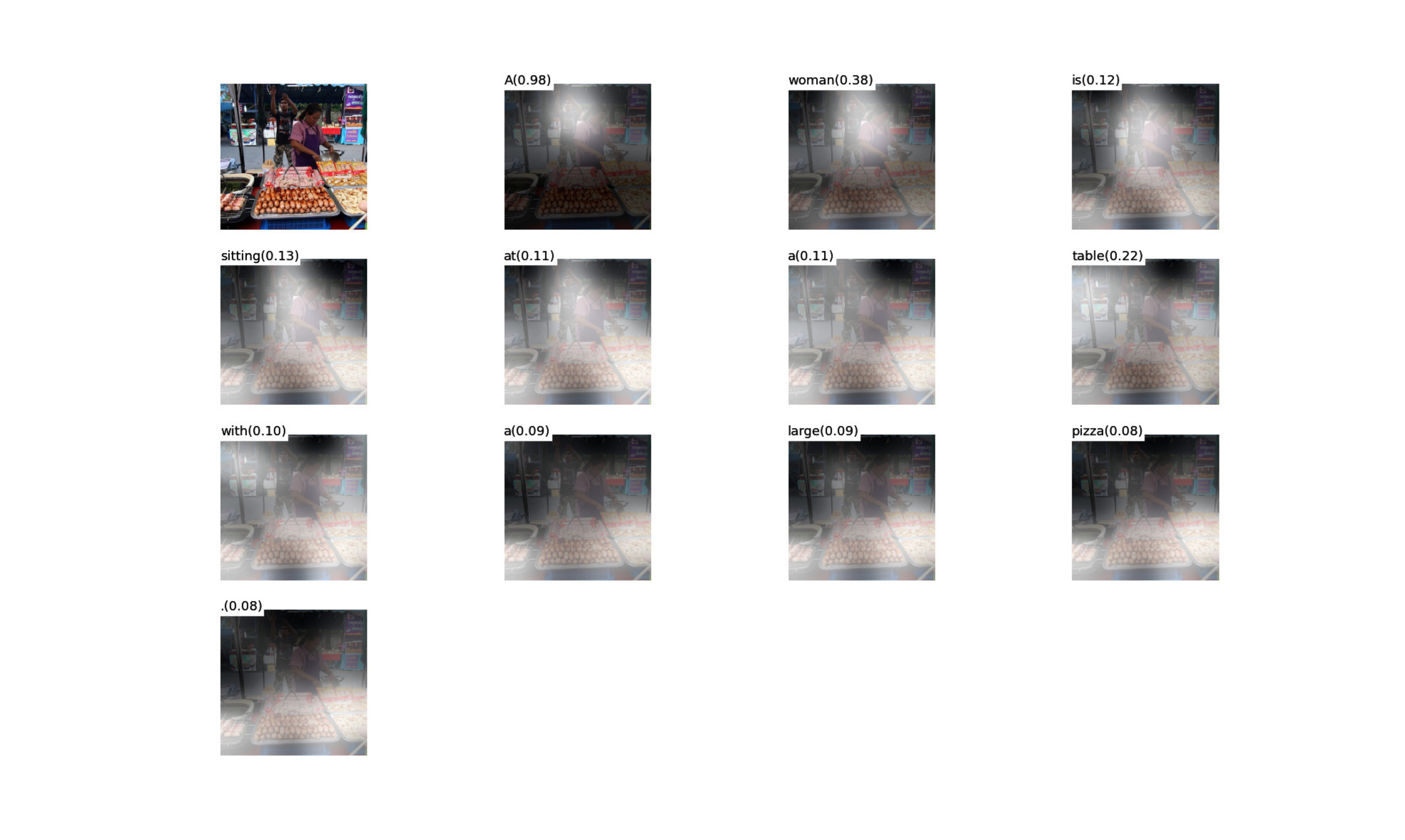}}
\vskip -0.4in
(b) A woman is sitting at a table with a large pizza.
\caption{}
\label{figure:im462_1252}
\end{center}
\vskip -0.4in
\end{figure*}

\begin{figure*}[ht]
\begin{center}
\centerline{\includegraphics[width=7.3in]{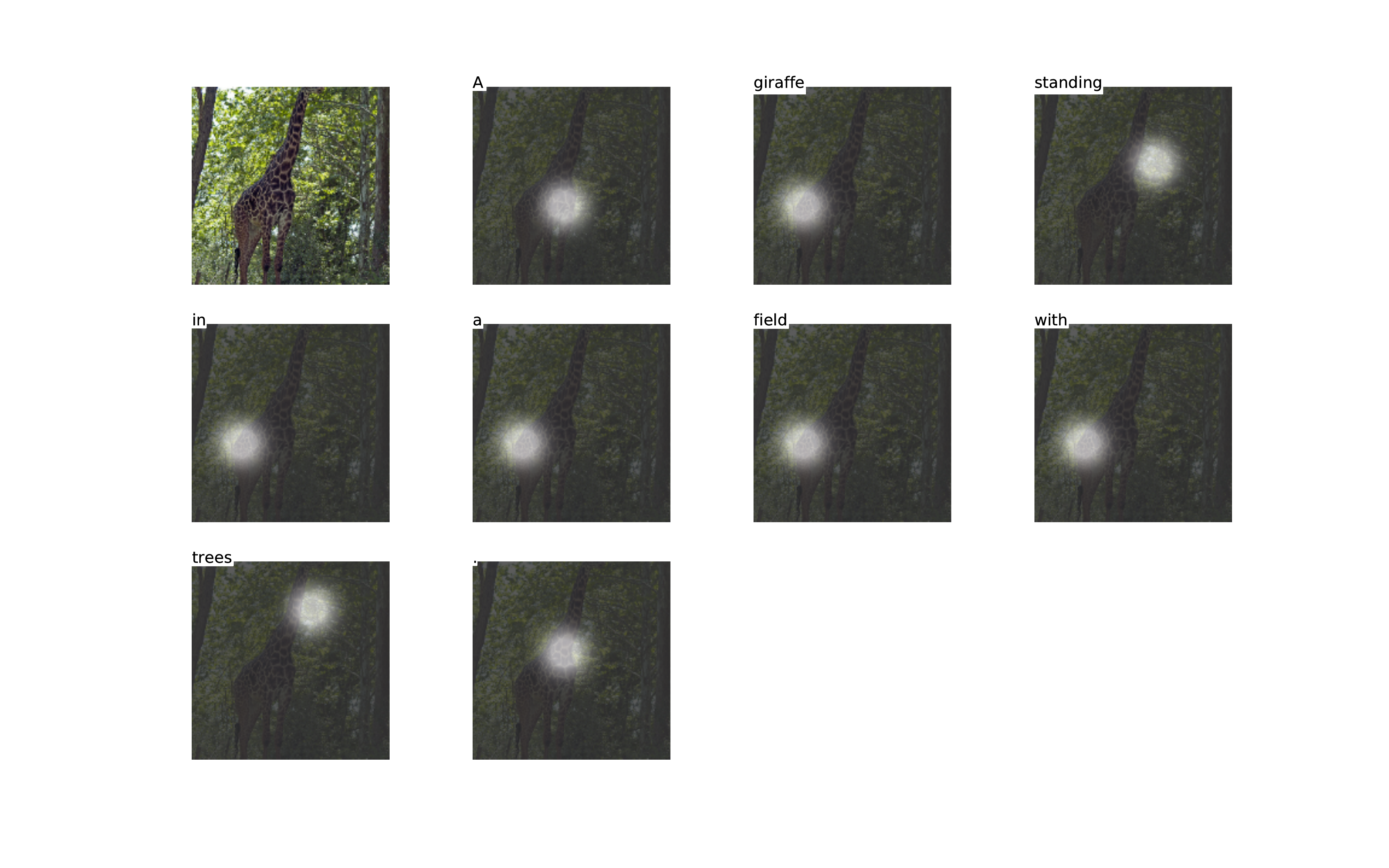}}
\vskip -0.3in
(a) A giraffe standing in a field with trees.
\centerline{\includegraphics[width=7.3in]{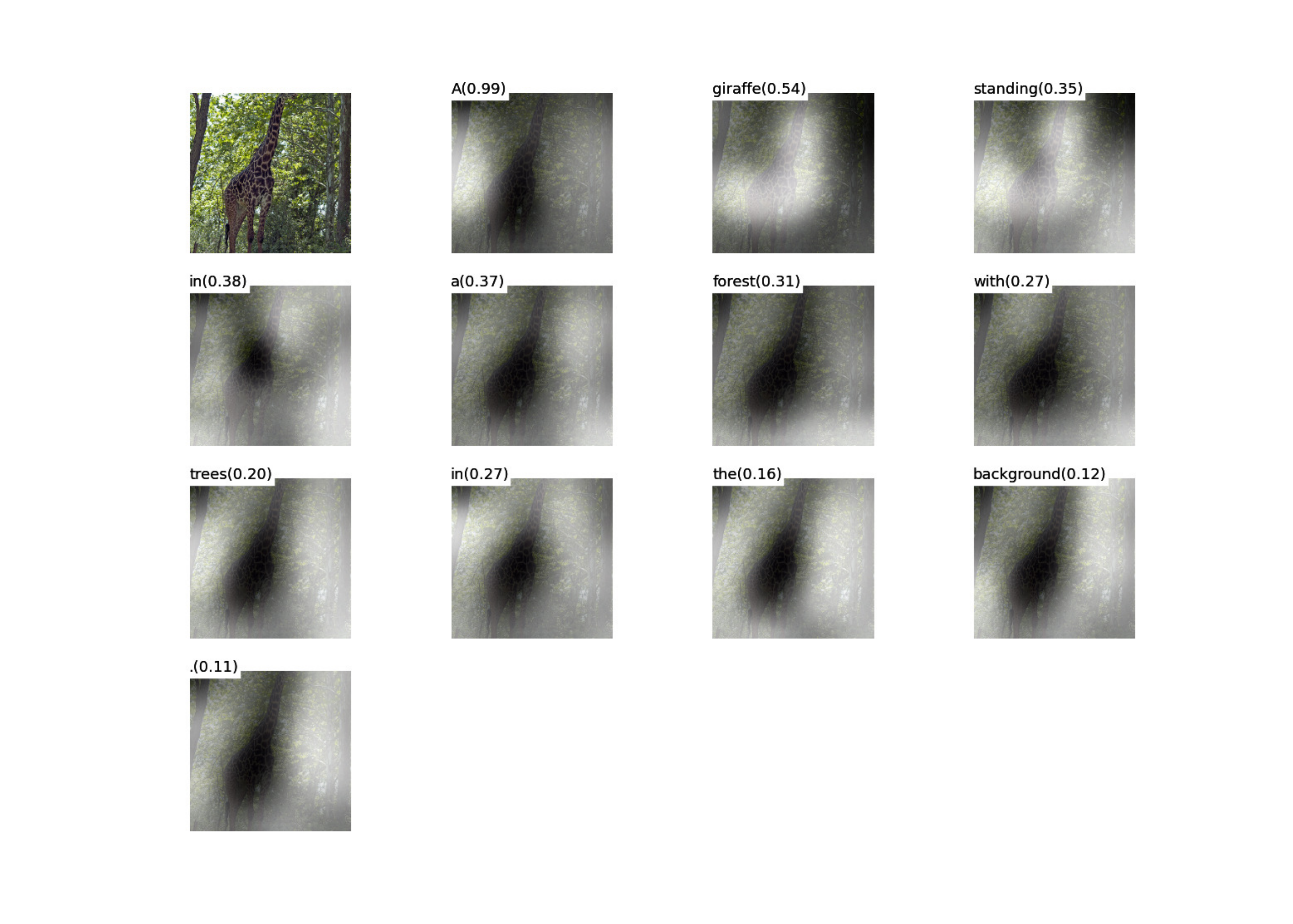}}
\vskip -0.4in
(b) A giraffe standing in a forest with trees in the background.
\caption{}
\label{figure:im1300_4837}
\end{center}
\vskip -0.4in
\end{figure*}

\begin{figure*}[ht]
\begin{center}
\centerline{\includegraphics[width=7.5in]{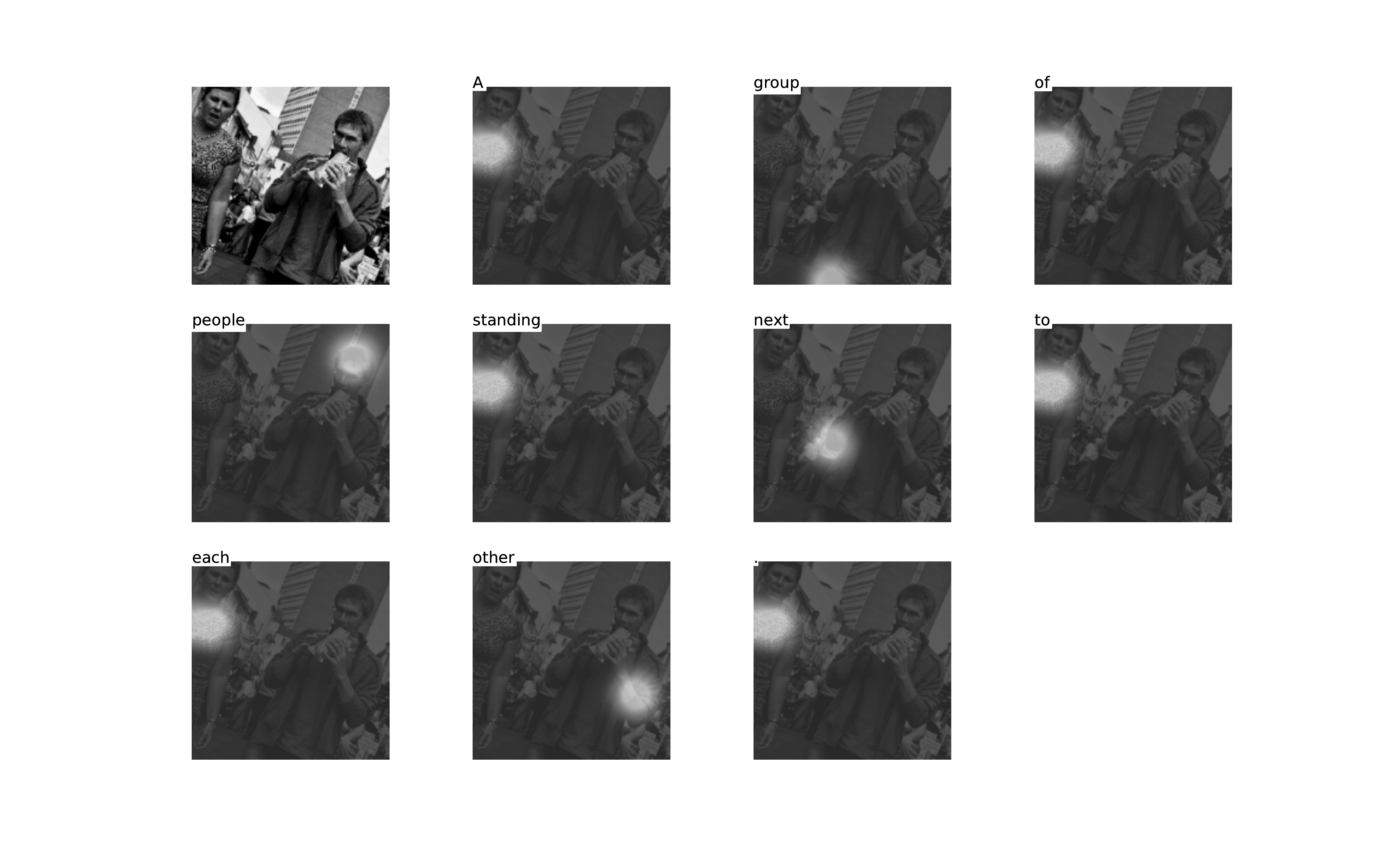}}
\vskip -0.2in
(a) A group of people standing next to each other.
\centerline{\includegraphics[width=7.5in]{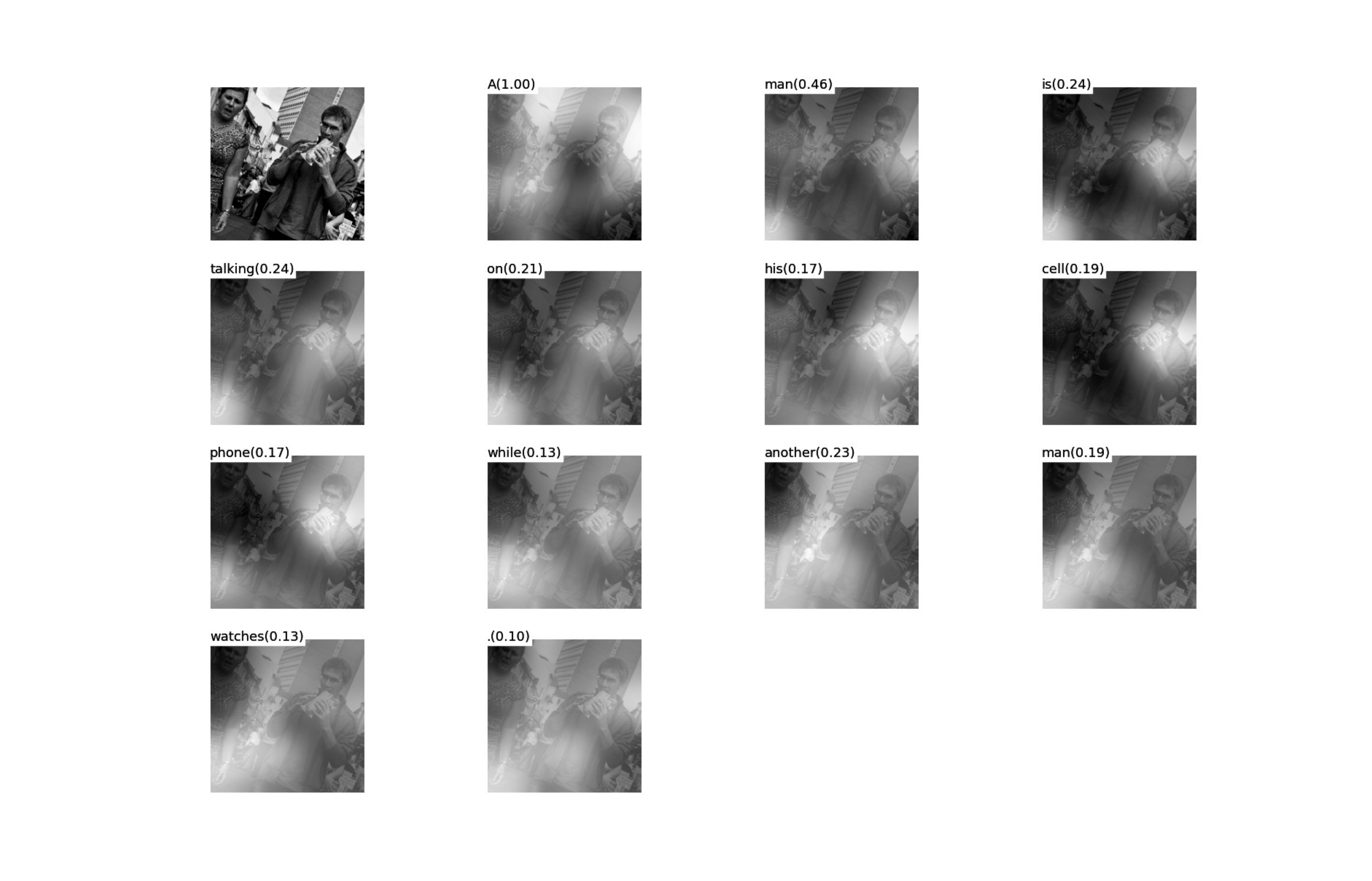}}
\vskip -0.4in
(b) A man is talking on his cell phone while another man watches.
\caption{}
\label{figure:im1228}
\end{center}
\vskip -0.4in
\end{figure*}